\documentclass[10pt,twocolumn,letterpaper]{article}
\usepackage[pagenumbers]{cvpr}
\usepackage{amsmath,amssymb}
\usepackage{graphicx}
\usepackage{booktabs}
\usepackage{colortbl}
\usepackage{caption}
\usepackage{xspace}
\definecolor{cvprblue}{rgb}{0.21,0.49,0.74}
\definecolor{retouchpale}{HTML}{F8E1EB}
\definecolor{retouchdeep}{HTML}{A62F63}
\usepackage[pagebackref,breaklinks,colorlinks,allcolors=cvprblue]{hyperref}
\AtEndPreamble{%
    \crefname{figure}{Figure}{Figures}
    \Crefname{figure}{Figure}{Figures}
    \crefname{table}{Table}{Tables}
    \Crefname{table}{Table}{Tables}
    \crefname{section}{Section}{Sections}
    \Crefname{section}{Section}{Sections}
    \crefname{subsection}{Section}{Sections}
    \Crefname{subsection}{Section}{Sections}
    \crefname{subsubsection}{Section}{Sections}
    \Crefname{subsubsection}{Section}{Sections}
    \crefname{appendix}{Appendix}{Appendices}
    \Crefname{appendix}{Appendix}{Appendices}
}
\newcommand{\method}{ReTouchGS\xspace}
\newcommand{\bench}{ReTouch3D\xspace}
\newcommand{\dE}{\ensuremath{\Delta E_{76}}\xspace}

\title{\method: Ultralight Render-free Aesthetic Color Enhancement and Transfer for 3D Gaussian Splatting}
\author{%
    Chuanzhi Xu\textsuperscript{1,*,$\dagger$}\quad
    Boyu Wei\textsuperscript{1,*}\quad
    Chengkun Yue\textsuperscript{2}\quad
    Xuanhua Yin\textsuperscript{1}\\
    Yiyang Li\textsuperscript{1}\quad
    Haodong Chen\textsuperscript{1}\quad
    Qiang Qu\textsuperscript{1}\quad
    Weidong Cai\textsuperscript{1}\\[3pt]
    {\normalsize
        \textsuperscript{1}The University of Sydney\quad
        \textsuperscript{2}Indiana University\quad
        \href{https://github.com/LouckXu/ReTouchGS}{%
            \raisebox{-0.12em}{%
                \includegraphics[height=0.9em]{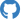}%
            }\hspace{0.3em}Project Page%
        }%
    }%
}
\hypersetup{%
    pdftitle={ReTouchGS: Ultralight Render-free Aesthetic Color Enhancement and Transfer for 3D Gaussian Splatting},
    pdfauthor={Chuanzhi Xu, Boyu Wei, Chengkun Yue, Xuanhua Yin, Yiyang Li, Haodong Chen, Qiang Qu, Weidong Cai},
    pdfsubject={Aesthetic color enhancement and reference-guided color transfer for 3D Gaussian Splatting},
    pdfkeywords={3D Gaussian Splatting, aesthetic enhancement, color transfer, LUT}
}
\makeatletter
\renewcommand{\@maketitle}{%
    \begin{center}
        {\Large\bfseries\@title\par}
        \vspace{8pt}
        {\large\@author\par}
    \end{center}
    \vspace{4pt}
}
\makeatother

\begin{document}
\raggedbottom
\AddToHookNext{shipout/after}{\flushbottom}
\twocolumn[{
\maketitle
\begin{minipage}{\textwidth}
    \centering
    \includegraphics[width=\textwidth]{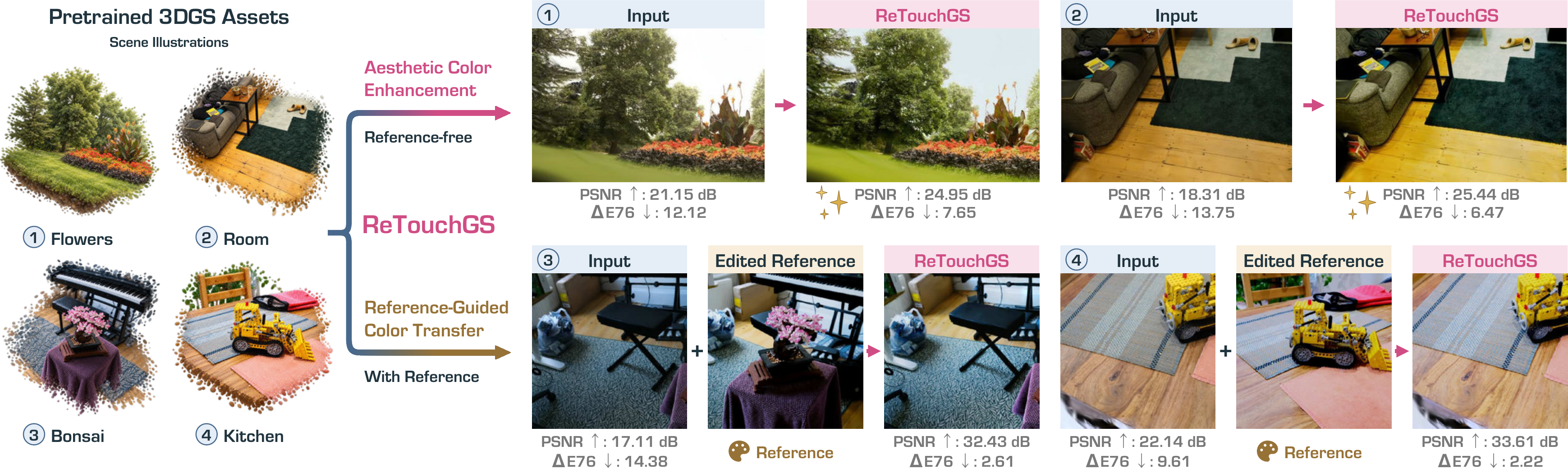}
    \captionsetup{hypcap=false}
    \captionof{figure}{\method enables automatic aesthetic color enhancement and reference-guided transfer with consistent colors across views.}
    \label{fig:teaser}
    \captionsetup{hypcap=true}
\end{minipage}
\vspace{12pt}
}]

\begingroup
\renewcommand{\thefootnote}{\fnsymbol{footnote}}
\footnotetext[1]{Equal contribution.}
\footnotetext[2]{Corresponding author:
    \href{mailto:chuanzhi.xu@sydney.edu.au}{%
        \texttt{chuanzhi.xu@sydney.edu.au}%
    }.}
\endgroup

\begin{abstract}
3D Gaussian Splatting (3DGS) enables photorealistic scene reconstruction.
Existing 3DGS color transfer and interactive editing methods are guided by reference images and user-specified color adjustments.
However, in many applications, users simply want aesthetically guided color enhancement that automatically makes the scene more visually appealing while maintaining consistent colors and tones across viewpoints.
We present \method, to our knowledge the first framework for reference-free aesthetic color enhancement of pretrained 3DGS models.
It predicts a scene-level color lookup table (LUT) from the model and its associated camera parameters.
\method also supports reference-guided color transfer by fitting a scene-level LUT using one edited view of the same scene and its corresponding original render.
A shared blend-aware color refinement network with only 44.7K parameters refines an analytic color update to realize either color mapping without rendering.
It preserves geometry and delivers standard 3DGS assets in seconds.
We also design the \bench benchmark to support the evaluation of aesthetic color enhancement and reference-guided color transfer across viewpoints.
Experiments demonstrate high color accuracy and strong generalization to unseen scenes in both aesthetic color enhancement and reference-guided color transfer.
\end{abstract}

\section{Introduction}
\label{sec:intro}

3D Gaussian Splatting (3DGS)~\cite{kerbl3Dgaussians} enables photorealistic reconstruction and real-time rendering of reusable scene assets.
As these assets support immersive viewing and visual creation, their aesthetic appeal becomes increasingly important.
Adjusting a scene's exposure, color balance, and tone can improve its overall aesthetic presentation.
Therefore, creators need tools that automatically enhance scene colors or transfer preferred colors and tones from an edited view, with consistent results across viewpoints (\cref{fig:teaser}).

In 2D photography, learning from expert retouching provides a foundation for automatic aesthetic enhancement.
Datasets such as MIT-Adobe FiveK~\cite{bychkovsky2011fivek} and PPR10K~\cite{liang2021ppr10k} pair photographs with expert edits, making photographic aesthetic preferences available for learning.
Image enhancement methods learn these preferences through image-adaptive lookup tables (LUTs)~\cite{zeng2020learning,yang2022adaint,yang2022seplut} and color-aware tone curves~\cite{serrano2024namedcurves}, adjusting color and tone based on image content to improve aesthetic presentation.
Users may also want to bring these aesthetic enhancements to 3D scenes.

Several studies support 3D appearance editing guided by reference images or user input.
Reference-guided color transfer~\cite{wang2024bilarf,xiao2025multiview} propagates adjustments from a retouched view, and interactive editing~\cite{reparamgs} provides palette and luminance controls.
Similarly, reference-based stylization~\cite{fpgs,zhuang2025styleme3d} transfers a chosen visual style.
However, users often seek a more visually appealing scene without having a reference image or a specific set of color adjustments in mind.
Meeting this need requires automatic, reference-free aesthetic color enhancement of pretrained 3DGS assets, a task that existing work has not addressed.

Extending aesthetic enhancement to 3DGS presents three challenges.
First, image-adaptive enhancement applied independently to each view can produce different colors and tones~\cite{xiao2025multiview}, requiring a consistent retouching decision for the whole scene.
Second, preserving the retouched appearance in the model for subsequent novel-view rendering requires faithfully encoding this decision in Gaussian color parameters.
However, the order of nonlinear color transformation and Gaussian blending affects the resulting pixel colors, so directly transforming each Gaussian's color produces discrepancies with the target pixel colors.
Third, rendering-based editing~\cite{wang2024bilarf,saad} repeatedly renders the scene during per-scene optimization, adding substantial computation even for color adjustments.
Such optimization must also preserve the reconstructed geometry and view-dependent appearance to avoid structural or appearance artifacts.
These requirements motivate render-free color updates that preserve geometry and opacity.

In this paper, we propose \method, to our knowledge the first framework for reference-free aesthetic color enhancement of pretrained 3DGS assets, and it also supports reference-guided color transfer.
For reference-free enhancement, an aesthetic enhancement network learns from an image enhancer trained on expert-retouched photographs and predicts a scene-level LUT from the asset and its associated camera parameters.
The resulting RGB-to-RGB mapping is shared across viewpoints of the 3DGS scene.
For color transfer, we instead fit this mapping to a user-edited view and its corresponding original render.
Both tasks use a shared render-free color update stage that combines an analytic color update with learned color corrections from the blend-aware color refinement network, preserving geometry and opacity in a standard 3DGS asset.
The framework uses only 180.2K network parameters for reference-free enhancement and 44.7K for color transfer.

We also introduce the \bench benchmark with scene-consistent color targets for 37 scenes.
\method delivers retouched assets in seconds and achieves mean color errors of 5.50 \dE for aesthetic enhancement and 4.03 \dE for reference-guided color transfer on views beyond the capture trajectory. It outperforms all baselines in color accuracy, PSNR, and LPIPS on both tasks.

Our contributions are summarized as follows:
\begin{itemize}
    \item We propose \method, the first framework, to our knowledge, for reference-free aesthetic color enhancement of pretrained 3DGS assets, with support for reference-guided color transfer.
    \item We introduce the \bench benchmark with scene-consistent color targets and off-trajectory evaluation of retouching fidelity and cross-view generalization.
    \item We develop a shared blend-aware color refinement network that refines analytic color updates to embed predicted or reference-fitted color mappings in standard Gaussian assets without rendering. Experiments show that \method outperforms all evaluated baselines in color accuracy, PSNR, and LPIPS on both tasks.
\end{itemize}

\begin{figure*}[!t]
    \centering
    \includegraphics[width=\textwidth]{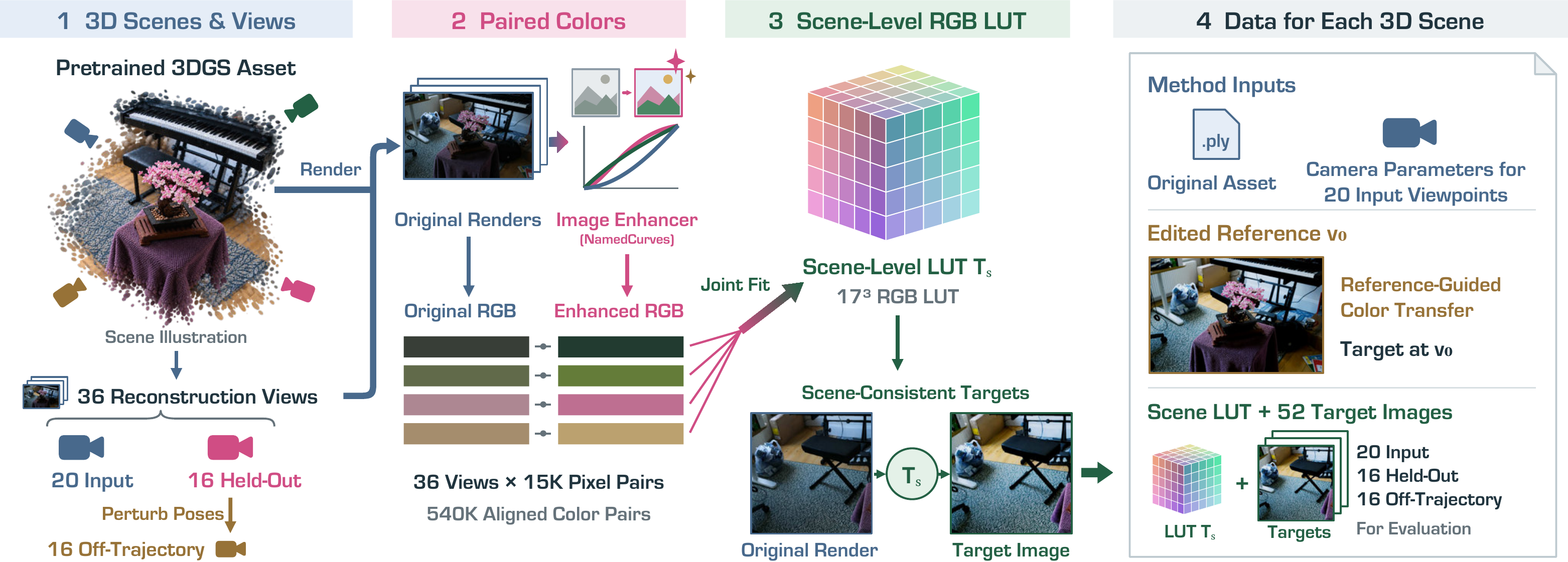}
    \caption{Data construction for each 3D scene in \bench. We jointly fit a scene-level RGB LUT to multi-view enhancement pairs to generate scene-consistent targets, including the edited reference for reference-guided color transfer.}
    \label{fig:benchmark}
\end{figure*}

\section{Related Work}
\label{sec:related}

\noindent{\textbf{2D Image Aesthetic Enhancement.}}
Expert-retouched datasets~\cite{bychkovsky2011fivek,liang2021ppr10k} support learning aesthetic color and tonal adjustments for 2D photographs.
Image-adaptive 3D LUTs~\cite{zeng2020learning} represent these adjustments by combining learned basis tables.
AdaInt~\cite{yang2022adaint} adapts their color-space sampling, while SepLUT~\cite{yang2022seplut} separates channel-independent adjustments from channel interactions.
NamedCurves~\cite{serrano2024namedcurves} learns tone curves for named colors and combines the adjusted images through attention.
These methods establish strong photographic retouching priors at the image level.
\method brings the learned aesthetic decision to a pretrained 3DGS asset, predicting scene-wide colors and tones and embedding them in Gaussian colors for consistent novel-view rendering.

\noindent{\textbf{3D Color Editing and Transfer.}}
BilaRF~\cite{wang2024bilarf} and Xiao et al.~\cite{xiao2025multiview} propagate enhancement from an edited view to neural radiance fields and 3DGS, respectively.
IReNe~\cite{mazzucchelli2024irene} and VIRGi~\cite{virgi} support user-directed recoloring while preserving view-dependent effects, and ReparamGS~\cite{reparamgs} provides interactive palette and luminance controls.
Instant Colorization~\cite{instantcolorization} and Support-Aware Appearance Distillation (SAAD)~\cite{saad} fit Gaussian colors to edited targets through visibility-weighted least squares and spherical harmonic distillation, respectively.
FPGS~\cite{fpgs} and StyleMe3D~\cite{zhuang2025styleme3d} transfer reference styles to Gaussian scenes, with StyleMe3D incorporating perceptual quality priors.
ViP3DE~\cite{chen2026vip3de} uses video priors for multi-view consistent editing.
Rendering-based optimization methods, such as BilaRF~\cite{wang2024bilarf} and SAAD~\cite{saad}, repeatedly render the scene during editing, adding computational overhead even for color adjustments.
However, these methods require users to specify the target appearance through reference examples or interactive controls.
Luminance-GS~\cite{cui2025luminancegs} and LITA-GS~\cite{zhou2025litags} address reconstruction under challenging illumination, restoring scene appearance from degraded observations.
In contrast, \method automatically determines how to improve a pretrained 3DGS scene's existing colors based on aesthetic criteria without a reference, and efficiently realizes that decision through render-free color updates that preserve geometry and opacity.
The same LUT representation and color refinement network also support reference-guided color transfer.

\section{\bench Benchmark}
\label{sec:bench}
\label{sec:benchmark_tasks}

Paired 3D scenes with view-consistent photographic retouching targets are scarce.
We construct \bench to support the evaluation of 3D aesthetic color enhancement and reference-guided color transfer using scene-consistent targets derived from expert retouching priors (\cref{fig:benchmark}).

The benchmark evaluates two tasks on 3D scenes: reference-free aesthetic color enhancement and reference-guided color transfer.
Reference-free aesthetic color enhancement takes a 3D asset $\mathcal G_s$ and its associated camera parameters $\mathcal C_s$, including intrinsics and poses, and predicts aesthetically enhanced colors and tones using learned photographic retouching priors.
Reference-guided color transfer additionally receives one edited view of the same scene at reference viewpoint $v_0$, to specify the preferred colors and tones.
In both tasks, ReTouchGS outputs a retouched standard 3DGS asset whose novel views express the desired colors and tones.
The two tasks share the same target colors and tones, along with the evaluation views.

\subsection{Aesthetic Target Construction}
\label{sec:benchmark_targets}

An image enhancer generates aesthetic targets from original renders at reconstruction viewpoints.
We use NamedCurves MIT5K-DPE~\cite{serrano2024namedcurves}, trained on expert edits from MIT-Adobe FiveK~\cite{bychkovsky2011fivek}.
The image enhancer independently adjusts colors and tones according to the content visible in each view, which can lead to cross-view inconsistencies.
Therefore, we jointly fit one scene-level RGB grid LUT to multiple enhanced reconstruction views to establish a consistent color mapping for the entire scene.
Pooling views exposes the fit to different scene regions and color distributions.
Let $T_s$ denote the fitted operator, $R$ the 3DGS renderer, and $\Pi$ clipping to $[0,1]$.
The target at any viewpoint $v$ is:
\begin{equation}
    I_s^\star(v)=T_s\!\left(\Pi(R(\mathcal G_s,v))\right).
    \label{eq:benchmark_target}
\end{equation}
This shared operator fixes the color mapping across viewpoints.
Targets are constructed offline, and the evaluator stores the scene LUT and all target images except the supplied edited reference image.
\Cref{app:benchmark} specifies sampling, fitting, and quantization.

\subsection{Evaluation Protocol}
\label{sec:benchmark_protocol}

\bench contains 28 test scenes from DL3DV~\cite{ling2024dl3dv} and all nine from Mip-NeRF 360~\cite{barron2022mipnerf360}.
Input and held-out reconstruction views are separated.
A reusable asset must also preserve the chosen colors and tones when visibility and Gaussian blending change beyond the capture trajectory.
Therefore, the primary evaluation uses 16 off-trajectory views per scene, measuring mean pixelwise \dE in CIELAB~\cite{cie2019cielab} against the scene-consistent targets.
Scores first average views within a scene, then weight all 37 scenes equally.
\Cref{app:benchmark} provides scene statistics, evaluation-view construction, and metric definitions.

\begin{figure*}[!t]
    \centering
    \includegraphics[width=\textwidth]{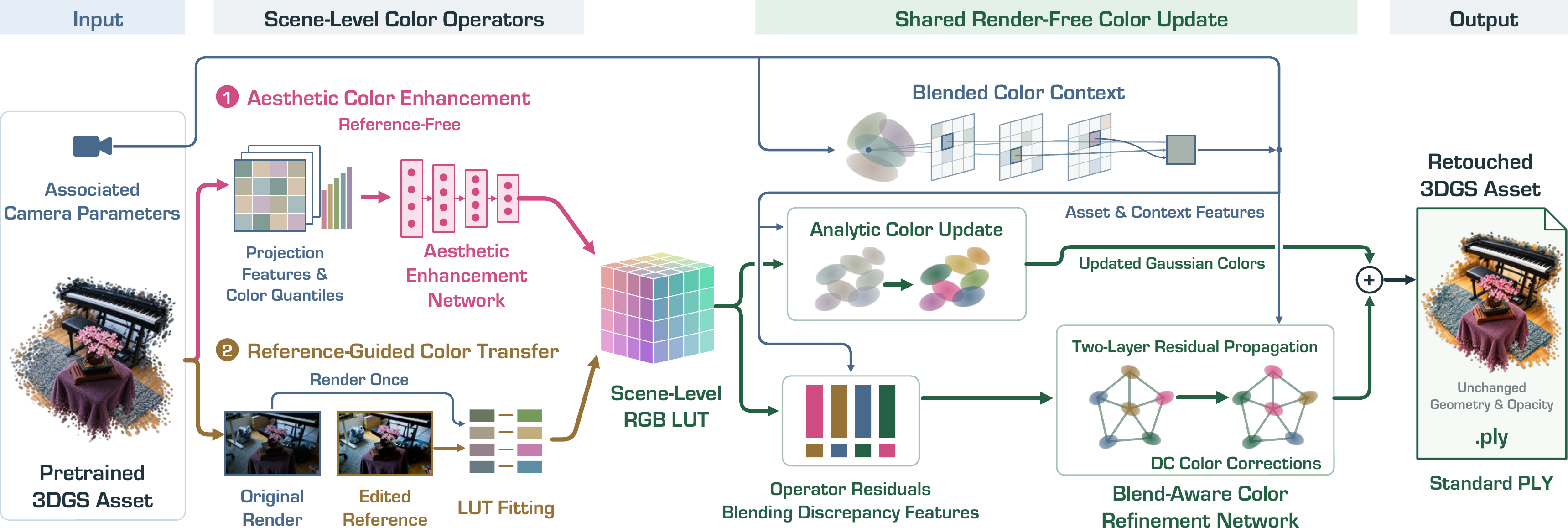}
    \caption{Overview of \method. Aesthetic color enhancement predicts a scene-level RGB LUT. Reference-guided color transfer fits a scene-level RGB LUT using an original render and its edited reference. Both tasks share a render-free color update stage that combines analytic color updates with learned color corrections, producing a standard 3DGS asset with fixed geometry and opacity.}
    \label{fig:method_overview}
\end{figure*}

\section{Methodology - \method}
\label{sec:method}

\method separates the choice of a scene's colors and tones from their realization in Gaussian color parameters (\cref{fig:method_overview}).
Given an asset $\mathcal G$ and its associated camera parameters $\mathcal C$, an aesthetic enhancement network predicts a scene-level color operator for reference-free enhancement.
For reference-guided color transfer, the operator is fitted to an edited view and its original render.
Both operators enter a shared render-free color update stage.
An analytic color update provides an initial transformation, and a blend-aware color refinement network predicts color corrections to produce an edited asset $\mathcal G'$.
This separation lets learned aesthetic preferences and user-specified colors and tones share one mechanism for consistent 3D retouching.

\subsection{Scene-Level Color Operators}
\label{sec:method_operator}

A scene-level color operator should apply the same photographic adjustment to corresponding colors across views.
We represent this adjustment by a color operator $T_\Theta$.
We use grid LUTs by default, predicted for enhancement and fitted to a reference pair for transfer.
A uniform RGB grid stores displacements from identity:
\begin{equation}
    T_\Theta(x)=\Pi\!\left(x+w(x)^\top\Theta\right),
    \label{eq:operator}
\end{equation}
where $\Theta\in\mathbb R^{D\times3}$ stores $D$ LUT-node displacements, $w(x)$ contains trilinear interpolation weights, and $\Pi$ clamps RGB values to $[0,1]$.
This representation couples color and tonal adjustments in one mapping and provides an explicit zero-edit state at $\Theta=0$.

\noindent{\textbf{Reference-Free Aesthetic Color Enhancement.}}
Choosing an aesthetic adjustment requires both the spatial appearance of a scene and its overall color distribution.
Our aesthetic enhancement network $E_\phi$ obtains these complementary cues by projecting Gaussian centers into coarse image-plane projection grids.
It encodes view-dependent spherical harmonic (SH) colors, opacity, and relative scales, then pools features with opacity and depth weights.
A convolutional encoder summarizes the resulting spatial features, while RGB and luma quantiles describe the projected color and tonal distribution.
A feature encoder combines these cues at each viewpoint, and mean pooling across viewpoints forms one scene descriptor.

Expert-derived scene LUTs contain recurring, correlated photographic adjustments.
Therefore, principal component analysis (PCA) of the training LUTs defines a compact space for predicting these adjustments.
The network predicts standardized coefficients and reconstructs the LUT as:
\begin{equation}
    \begin{aligned}
        z_\phi&=E_\phi(\mathcal G,\mathcal C),\\
        \operatorname{vec}(\widehat\Theta)&=\mu+B(\sigma\odot z_\phi),
    \end{aligned}
    \label{eq:enhancement_pca}
\end{equation}
where $\mu$ is the mean training LUT, the columns of $B$ are its principal directions, and $\sigma$ contains coefficient standard deviations.
\Cref{app:implementation} details the projection and feature construction.

\noindent{\textbf{Reference-Guided Color Transfer.}}
An edited view specifies the creator's preferred colors and tones through corresponding original and target colors.
The original asset is rendered once from the reference viewpoint, and a grid LUT is fitted to the paired original and edited RGB values.
An identity-centered ridge penalty anchors colors sparsely represented in the reference, allowing one edited view to specify a scene-level mapping.
The fitted LUT enters the shared render-free color update stage.
The fitting objective and solver settings are specified in \cref{app:implementation}.

\subsection{Blended Color Context and Analytic Color Update}
\label{sec:method_context}

With geometry and opacity fixed, rendered pixels combine contributions from several Gaussians.
However, applying a nonlinear operator to this mixture differs from mixing individually transformed colors.
Therefore, each Gaussian's update should account for the colors it contributes to forming.
The blended color context $m_i$ is estimated by pooling view-dependent SH colors in coarse projection grids with opacity and soft depth weights, then aggregating the grid colors at Gaussian $i$'s projections across viewpoints.
This supplies a color around which to approximate the input color transformation using asset attributes and camera projections.

Let $c_i$ be Gaussian $i$'s view-independent (DC) color represented by the zeroth-order SH coefficient, $h_{i\ell}$ its higher-order SH coefficients, and $u_i=\Pi(c_i)$.
A first-order expansion of $T$ around $m_i$ uses the Jacobian $J_T(m_i)$.
The context weight $a_i$ decreases as the Gaussian's own color departs from its context, transitioning toward an update at $u_i$.
The resulting analytic color update is:
\begin{equation}
    \begin{aligned}
        a_i&=\exp\!\left(-\frac{\|c_i-m_i\|_2^2}{\delta^2}\right),\\
        c_i^{\mathrm{cf}}
        &=a_i\bigl[T(m_i)+J_T(m_i)(c_i-m_i)\bigr]\\
        &\quad +(1-a_i)\bigl[T(u_i)+c_i-u_i\bigr],\\
        h_{i\ell}^{\mathrm{cf}}
        &=\bigl[a_iJ_T(m_i)+(1-a_i)J_T(u_i)\bigr]h_{i\ell},
    \end{aligned}
    \label{eq:analytic_update}
\end{equation}
where $\ell>0$ and $\delta$ controls the decay of the context weight.
The DC update changes the base color, while the local Jacobian transforms higher-order SH coefficients so that view-dependent variations follow the same adjustment.
The term $c_i-u_i$ preserves the DC component outside the LUT domain.

\subsection{Blend-Aware Color Refinement Network}
\label{sec:method_writer}

The analytic color update captures a local first-order transformation.
To account for operator curvature and the approximation of blended color context, the network learns a DC correction through residual propagation between nearby Gaussians.
This formulation makes the desired color changes explicit and lets the network learn how to distribute them.

\noindent{\textbf{Operator Residuals and Blending Discrepancy Features.}}
Three operator residuals describe direct color displacement, context displacement, and the linearized change in the Gaussian-to-context color difference:
\begin{equation}
    \begin{aligned}
        r_i^{\mathrm{dir}}&=T(u_i)-u_i,\\
        r_i^{\mathrm{ctx}}&=T(m_i)-m_i,\\
        r_i^{\mathrm{jac}}&=\widetilde{J_T(m_i)-I}\,(\tilde c_i-m_i).
    \end{aligned}
    \label{eq:operator_residuals}
\end{equation}
The matrix $I$ is the $3\times3$ identity, and the tildes denote bounded features used in the learned correction, with their ranges specified in \cref{app:writeback}.
To expose the mixing discrepancy directly, a blending discrepancy feature is defined as:
\begin{equation}
    b_T(a,b)=T\!\left(\frac{a+b}{2}\right)-\frac{T(a)+T(b)}{2}.
    \label{eq:blend_error}
\end{equation}
This blending discrepancy feature measures the difference between transforming a mixture and mixing its transformed colors.
Each node uses $b_T(u_i,m_i)$, and each edge uses $b_T(u_i,u_j)$.
These LUT queries expose color-space curvature beyond the analytic color update.

\noindent{\textbf{Residual Propagation.}}
A 3D nearest-neighbor graph propagates color corrections across local Gaussian neighborhoods.
Node and edge features encode Gaussian appearance, blended color context, and relative geometry.
These asset-derived features predict scalar propagation coefficients shared by the three RGB channels.
The input color operator supplies the residual channels, allowing the same network to handle different color and tonal adjustments.

The network has two residual-propagation layers.
The second layer appends first-layer corrections to its node and neighbor residuals, allowing neighboring corrections to inform the final update.
Let $\mathcal N^+(i)$ contain node $i$ and its neighbors, and let $x_{ij,k}^{(h)}$ collect the applicable residuals and blending discrepancy features at layer $h$.
With learned coefficients $\alpha_{ijk}^{(h)}$, the color update gives:
\begin{equation}
    \begin{aligned}
        d_{\theta,i}&=\sum_{h=1}^{2}\sum_{j\in\mathcal N^+(i)}\sum_k
        \alpha_{ijk}^{(h)}x_{ij,k}^{(h)},\\
        c_i'&=c_i^{\mathrm{cf}}+d_{\theta,i},\qquad
        h_{i\ell}'=h_{i\ell}^{\mathrm{cf}}.
    \end{aligned}
    \label{eq:writeback}
\end{equation}
For the identity operator, all residuals and blending discrepancy features vanish, making both learned layers output zero corrections.
Together with the analytic color update, this preserves the original colors by construction.
The updated SH coefficients form a standard 3DGS asset with the original geometry and opacity.

\begin{figure*}[!t]
\centering
\includegraphics[width=\textwidth]{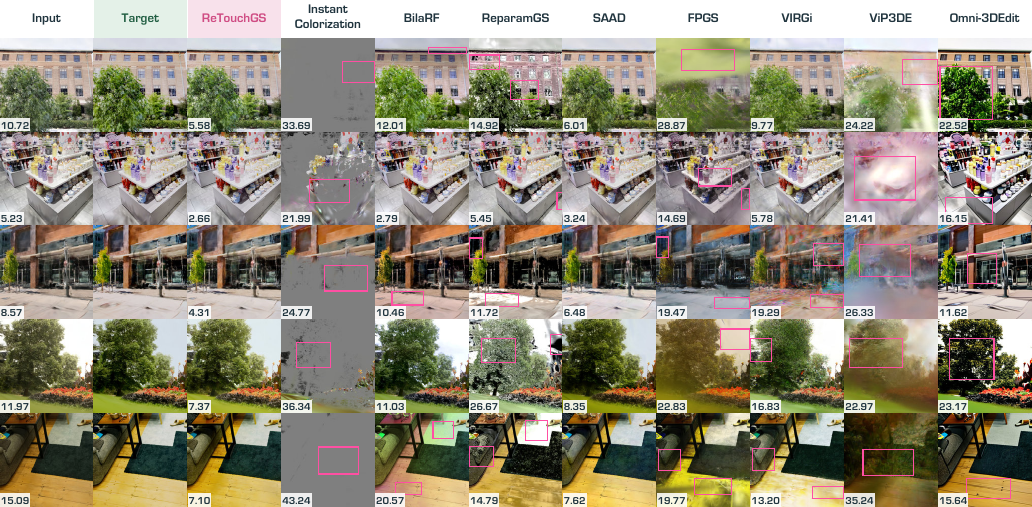}
\caption{Qualitative results of aesthetic color enhancement. Outputs in each row share the same off-trajectory viewpoint. Pink boxes highlight color artifacts in the enhanced results. The annotated values give the per-view \dE$\downarrow$ color error against the target images.}
\label{fig:exp_qualitative}
\end{figure*}

\subsection{Training Strategy and Objectives}
\label{sec:method_training}

\noindent{\textbf{Learning Aesthetic Decisions.}}
The aesthetic enhancement network learns scene-level target operators $T_s$ constructed from training scenes using NamedCurves (\cref{sec:benchmark_targets}).
To supervise the colors that determine each asset's appearance, a color bank $\mathcal X_s$ samples colors from Gaussian--viewpoint pairs according to projected footprint.
The objective is:
\begin{equation}
    \mathcal L_{\mathrm{enh}}=\frac{1}{3}\mathbb E_{s,x\sim\mathcal X_s}
    \|T_{\widehat\Theta_s}(x)-Q_8(T_s(x))\|_1,
    \label{eq:enhancement_loss}
\end{equation}
where $Q_8$ denotes 8-bit quantization.
Training aggregates features from random input-viewpoint subsets, while inference uses all input viewpoints.

\noindent{\textbf{Learning Color Refinement.}}
The color refinement network learns across scenes and target operators, with LUT interpolation and displacement scaling extending the range of photographic adjustments.
Training proceeds from Gaussian-level supervision to image-level refinement.
First, a rendering-based color solver initialized with the analytic color update supplies target DC corrections $d_i^\star$.
The distillation objective is:
\begin{equation}
    \mathcal L_{\mathrm{distill}}
    =\frac{\sum_i\eta_i\|d_{\theta,i}-d_i^\star\|_2^2}{\sum_i\eta_i},
    \label{eq:distill_loss}
\end{equation}
where $\eta_i$ weights accumulated visibility.
This teaches how Gaussians compensate for each other's contributions to a target image.

Next, rendering fine-tuning aligns the predicted corrections with their effect after Gaussian blending.
Reconstruction viewpoints and perturbed viewpoints expose the same edited asset to changing visibility during training.
For a sampled operator $T$, write the target in \cref{eq:benchmark_target} as $I^\star_{s,T}(v)=T(\Pi(R(\mathcal G_s,v)))$.
The objective is:
\begin{equation}
    \begin{aligned}
        \mathcal L_{\mathrm{render}}
        &=\mathbb E_{s,T,v}\bigl[\|R(\mathcal G_s',v)-I^\star_{s,T}(v)\|_1\bigr]\\
        &\quad+\lambda_{\mathrm{corr}}\operatorname{mean}_{i,\mathrm{RGB}}
        d_{\theta,i}^2,
    \end{aligned}
    \label{eq:render_loss}
\end{equation}
where the image loss averages pixels and channels, and $\lambda_{\mathrm{corr}}$ weights a penalty that keeps the learned correction close to the analytic starting point.
The two networks are trained separately, with Gaussian geometry and opacity fixed throughout.
The demonstration solver is detailed in \cref{app:solver_demonstrations}, and training configurations are provided in \cref{app:experimental_settings}.

\section{Experiments and Results}
\label{sec:experiments}

\subsection{Experimental Settings}
\label{sec:exp_settings}

\noindent{\textbf{Dataset and Metrics.}}
We evaluate both tasks on all 37 scenes in \bench.
The primary metric is off-trajectory \dE, which measures fidelity to the scene-consistent color target.
We also report PSNR, SSIM~\cite{wang2004ssim}, and LPIPS~\cite{zhang2018lpips} to assess image and perceptual fidelity.

\noindent{\textbf{Comparison Methods.}}
\Cref{tab:exp_main} lists the baselines used in our comparisons.
To adapt reference-guided baselines to aesthetic enhancement, we use NamedCurves MIT5K-DPE~\cite{serrano2024namedcurves} to generate target images from their required input views.
Each baseline then applies the adjustment through its native editing pipeline.
For reference-guided color transfer, all methods receive the same edited reference.
Input-view counts, native output representations, and baseline adaptations are detailed in \cref{app:experimental_settings}.

\begin{table*}[!t]
\centering
\footnotesize
\setlength{\tabcolsep}{0.9pt}
\begin{tabular*}{\textwidth}{@{\extracolsep{\fill}}l|l|cccccc|cccccc@{}}
\toprule
& & \multicolumn{6}{c|}{Aesthetic Color Enhancement} & \multicolumn{6}{c}{Reference-Guided Color Transfer} \\
\cmidrule(lr){3-8}\cmidrule(lr){9-14}
Method & Venue / Year & \dE$\downarrow$ & PSNR$\uparrow$ & SSIM$\uparrow$ & LPIPS$\downarrow$ & Time (s)$\downarrow$ & Params. & \dE$\downarrow$ & PSNR$\uparrow$ & SSIM$\uparrow$ & LPIPS$\downarrow$ & Time (s)$\downarrow$ & Params. \\
\midrule
\rowcolor[gray]{0.90}
\multicolumn{14}{l}{\textit{Per-scene Solving and Optimization}} \\
Instant Colorization~\cite{instantcolorization} & CVPR Findings, 2026 & 23.50 & 13.80 & 0.5784 & 0.4911 & 10.95 & 0.9936M & 23.35 & 13.79 & 0.5793 & 0.4883 & 8.98 & 0 \\
BilaRF~\cite{wang2024bilarf} & SIGGRAPH, 2024 & 6.30 & 26.97 & \textcolor{retouchdeep}{\textbf{0.9350}} & \underline{0.0679} & 122.54 & 0.9940M & \underline{5.73} & \underline{27.87} & \textcolor{retouchdeep}{\textbf{0.9360}} & \underline{0.0655} & 121.32 & 381 \\
SAAD~\cite{saad} & arXiv, 2026 & \underline{6.18} & \underline{27.05} & 0.9079 & 0.0847 & 3461.78 & 1.008M & 15.46 & 20.10 & 0.8249 & 0.2021 & 340.57 & 14.56K \\
ReparamGS~\cite{reparamgs} & arXiv, 2026 & 11.65 & 19.16 & 0.6933 & 0.2701 & 136.08 & 0.9936M & 11.44 & 19.40 & 0.7023 & 0.2658 & 6.02 & 0 \\
VIRGi~\cite{virgi} & TPAMI, 2026 & 11.88 & 20.11 & 0.7459 & 0.2186 & 22993.69 & 13.68M & 11.65 & 20.26 & 0.7467 & 0.2172 & 76.60 & 12.68M \\
\rowcolor[gray]{0.90}
\multicolumn{14}{l}{\textit{Learned Style and Edit Propagation}} \\
FPGS~\cite{fpgs} & IJCV, 2026 & 21.42 & 14.73 & 0.6260 & 0.3496 & 633.82 & 31.30M & 21.63 & 14.69 & 0.6249 & 0.3516 & 17.41 & 30.30M \\
ViP3DE~\cite{chen2026vip3de} & AAAI, 2026 & 22.21 & 14.31 & 0.5313 & 0.5748 & 279.26 & 3040M & 22.39 & 14.23 & 0.5283 & 0.5804 & 277.93 & 3039M \\
Omni-3DEdit~\cite{chen2026omni3dedit} & CVPR, 2026 & 14.37 & 17.57 & 0.6288 & 0.3069 & 173.92 & 4868M & 14.30 & 17.60 & 0.6311 & 0.3050 & 153.98 & 4867M \\
\midrule
\rowcolor{retouchpale}
\method & -- & \textcolor{retouchdeep}{\textbf{5.50}} & \textcolor{retouchdeep}{\textbf{28.08}} & \underline{0.9213} & \textcolor{retouchdeep}{\textbf{0.0561}} & 22.97 & 180.2K & \textcolor{retouchdeep}{\textbf{4.03}} & \textcolor{retouchdeep}{\textbf{29.49}} & \underline{0.9221} & \textcolor{retouchdeep}{\textbf{0.0529}} & 19.32 & 44.72K \\
\bottomrule
\end{tabular*}
\caption{Results of aesthetic color enhancement and reference-guided color transfer on \bench, averaged over 37 scenes. Params. counts network parameters. Bold deep pink marks the best quality scores, and underlining marks the second best.}
\label{tab:exp_main}
\end{table*}

\begin{table}[!t]
\centering
\footnotesize
\setlength{\tabcolsep}{1.5pt}
\resizebox{\columnwidth}{!}{%
\begin{tabular}{@{}l|cc|cc@{}}
\toprule
& \multicolumn{2}{c|}{Enhancement} & \multicolumn{2}{c}{Transfer} \\
\cmidrule(lr){2-3}\cmidrule(lr){4-5}
Configuration & \dE$\downarrow$ & PSNR$\uparrow$ & \dE$\downarrow$ & PSNR$\uparrow$ \\
\midrule
\multicolumn{5}{l}{\cellcolor[gray]{0.90}\textit{Color update components}} \\
w/o blended color context \& refinement & 6.417 & 26.66 & 5.299 & 27.76 \\
w/o blend-aware color refinement & 5.911 & 27.37 & 4.649 & 28.40 \\
\multicolumn{5}{l}{\cellcolor[gray]{0.90}\textit{Refinement mechanisms}} \\
w/o residual propagation & 5.594 & 27.93 & 4.189 & 29.21 \\
w/o blending discrepancy features & 5.563 & 28.01 & 4.147 & 29.33 \\
\midrule
\rowcolor{retouchpale}
Full model & \textcolor{retouchdeep}{\textbf{5.502}} & \textcolor{retouchdeep}{\textbf{28.08}} & \textcolor{retouchdeep}{\textbf{4.035}} & \textcolor{retouchdeep}{\textbf{29.49}} \\
\bottomrule
\end{tabular}%
}
\caption{Color update ablations averaged equally over 37 scenes.}
\label{tab:exp_ablation}
\end{table}

\noindent{\textbf{Implementation Details.}}
All experiments are conducted on a single NVIDIA RTX 5090 GPU.
All training stages use AdamW~\cite{loshchilov2019decoupled} with OneCycle learning-rate scheduling~\cite{smith2019superconvergence}.
First, we train the aesthetic enhancement network for 20,000 steps with a peak learning rate of $3\times10^{-4}$.
Next, we train the blend-aware color refinement network through solver distillation for 16,000 steps with a peak learning rate of $10^{-3}$.
Finally, we apply rendering fine-tuning to the refinement network for 12,000 steps with a peak learning rate of $3\times10^{-4}$.
Both tasks use a $17^3$ grid LUT, predicted for aesthetic enhancement and fitted to the edited reference for color transfer.
Full training and operator-fitting settings are in \cref{app:implementation,app:experimental_settings}.

\subsection{Comparative Results}
\label{sec:exp_comparison}

\noindent{\textbf{Aesthetic Color Enhancement.}}
\Cref{tab:exp_main} shows that \method achieves the lowest enhancement error while inferring aesthetically enhanced colors and tones directly from the asset, without running a 2D image enhancer at test time.
\Cref{fig:exp_qualitative} illustrates this fidelity in \textit{Flowers}' foliage and \textit{Room}'s warm floor and blue-green carpet, supporting asset-level prediction of photographic adjustments.

\begin{figure*}[t]
\centering
\includegraphics[width=\textwidth]{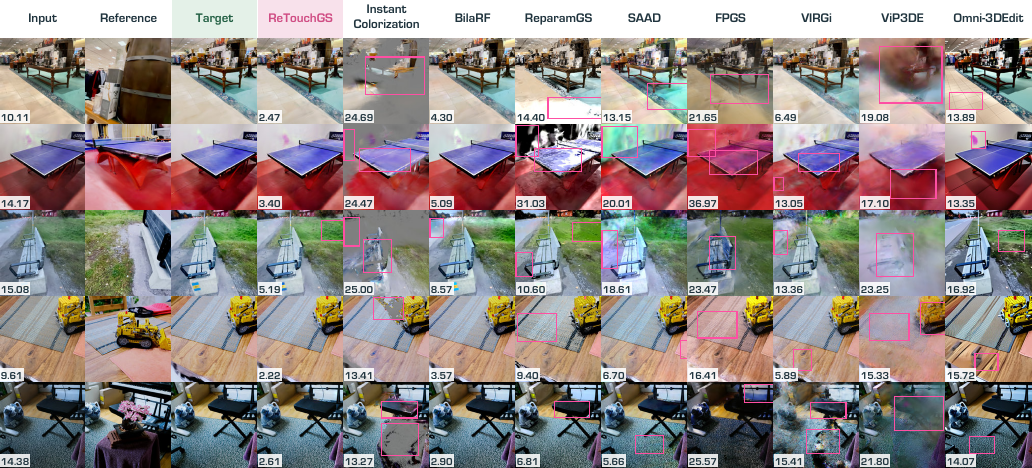}
\caption{Qualitative results of reference-guided color transfer. Reference shows the edited view used to specify the desired colors and tones. In each row, all outputs share the same off-trajectory viewpoint and crop. Pink boxes highlight color artifacts in the transferred results. The annotated values report the mean pixelwise \dE$\downarrow$ color error relative to Target over the shared crop.}
\label{fig:exp_qualitative_transfer}
\end{figure*}

\noindent{\textbf{Reference-Guided Color Transfer.}}
\Cref{tab:exp_main} shows that \method transfers reference colors and tones to novel views with the lowest color error and LPIPS and the highest PSNR among the compared methods.
\Cref{fig:exp_qualitative_transfer} shows preserved floor texture in \textit{Kitchen} and carpet tones in \textit{Bonsai}, illustrating how the shared color update combines control over appearance with scene detail preservation.

\begin{figure}[!t]
\centering
\includegraphics[width=\columnwidth]{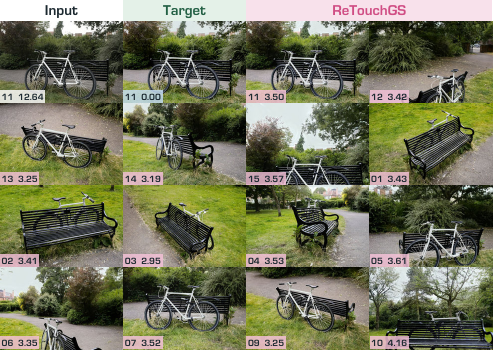}
\caption{Multi-view color transfer from one edited asset. Input and Target share a viewpoint. 14 held-out views follow. Corner labels give the view index and mean pixelwise \dE$\downarrow$ against the corresponding target.}
\label{fig:exp_multiview}
\end{figure}

\noindent{\textbf{View Generalization.}}
Held-out-view error measures color accuracy at reserved reconstruction viewpoints.
The view-generalization gap is the mean off-trajectory error minus the reference-view error.
\Cref{fig:exp_multiview} shows consistent foliage and lawn tones across 14 held-out views of the same edited \textit{Bicycle} asset.
More multi-view examples appear in \cref{app:extended_views}.
Reference-guided transfer gives similar errors of 3.81~\dE on the reference view and 3.85 on held-out views, averaged over all 37 scenes.
The small off-trajectory gaps further support consistent color adjustment across viewpoints, with detailed view-wise comparisons and temporal residuals in \cref{app:extended_results}.

\noindent{\textbf{Efficiency.}}
\Cref{tab:exp_main} shows that \method combines compact networks with efficient editing.
Its ultralight blend-aware color refinement network contains 44.7K parameters.
Aesthetic color enhancement uses feed-forward prediction, while reference-guided color transfer fits 14,739 LUT values with fixed refinement weights.
On an RTX 5090, median enhancement delivery takes 5.6 seconds for the DL3DV scenes and 32.8 seconds for the Mip-NeRF 360 scenes in \bench.
\method delivery times include asset loading, preparation, editing, and export.
Further timing and parameter details appear in \cref{app:experimental_settings,app:refinement_analysis}.

\begin{table}[!t]
    \centering
    \footnotesize
    \setlength{\tabcolsep}{1.4pt}
    \resizebox{\columnwidth}{!}{%
    \begin{tabular}{@{}l|cccc|cccc@{}}
        \toprule
        & \multicolumn{4}{c|}{Enhancement} & \multicolumn{4}{c}{Transfer} \\
        \cmidrule(lr){2-5}\cmidrule(lr){6-9}
        Method & \dE & PSNR & SSIM & LPIPS & \dE & PSNR & SSIM & LPIPS \\
        \midrule
        Instant Colorization~\cite{instantcolorization} & 23.89 & 13.61 & 0.5963 & 0.4920 & 23.62 & 13.66 & 0.5969 & 0.4915 \\
        BilaRF~\cite{wang2024bilarf} & 6.03 & 26.60 & 0.9521 & 0.0566 & 4.52 & 29.21 & 0.9598 & 0.0521 \\
        ReparamGS~\cite{reparamgs} & 10.78 & 19.25 & 0.7147 & 0.2476 & 10.06 & 19.58 & 0.7066 & 0.2528 \\
        Omni-3DEdit~\cite{chen2026omni3dedit} & 13.56 & 17.84 & 0.6304 & 0.2967 & 13.51 & 17.85 & 0.6305 & 0.2963 \\
        \midrule
        \rowcolor{retouchpale}
        \method & \textcolor{retouchdeep}{\textbf{5.13}} & \textcolor{retouchdeep}{\textbf{27.47}} & \textcolor{retouchdeep}{\textbf{0.9598}} & \textcolor{retouchdeep}{\textbf{0.0463}} & \textcolor{retouchdeep}{\textbf{3.08}} & \textcolor{retouchdeep}{\textbf{31.30}} & \textcolor{retouchdeep}{\textbf{0.9661}} & \textcolor{retouchdeep}{\textbf{0.0421}} \\
        \bottomrule
    \end{tabular}%
    }
    \caption{Aesthetic color enhancement and reference-guided color transfer with VeraRetouch targets on all 37 scenes in \bench.}
    \label{tab:app_vera}
\end{table}

\subsection{Ablation Studies}
\label{sec:exp_ablation}

\noindent{\textbf{Color Update Components.}}
As shown in \cref{tab:exp_ablation}, with the grid LUT fixed, accounting for blended color context improves over direct per-Gaussian transformation, and learned refinement further improves every test scene in reference-guided transfer.
This progression supports resolving how Gaussian colors blend when realizing the desired mapping.

\noindent{\textbf{Residual Propagation and Blending Discrepancy Features.}}
As shown in \cref{tab:exp_ablation}, removing either residual propagation or blending discrepancy features reduces color accuracy and PSNR in both tasks.
These results support propagating desired color changes through neighboring Gaussians and explicitly modeling their blending discrepancy.

\noindent{\textbf{Aesthetic Enhancement Network.}}
Combining projected appearance and color quantiles improves operator prediction over either input alone (\cref{app:refinement_analysis}).
The two inputs thus provide complementary cues for predicting aesthetic color adjustments.

\noindent{\textbf{Additional Analysis.}}
\Cref{app:refinement_analysis} shows that learned refinement improves color-mapping fidelity across seven LUT sources and that two-stage training outperforms rendering-only training.
The breakdowns by scene source also show consistent transfer gains in both groups.

\subsection{Adaptation to Different Retouching Targets}
\label{sec:exp_transfer}
To evaluate whether learned color refinement transfers to a new source of photographic adjustments, we use VeraRetouch~\cite{veraretouch} to construct targets for both aesthetic color enhancement and reference-guided color transfer on all 37 scenes in \bench.
The color refinement network was trained without color adjustments generated by VeraRetouch, and its weights remain fixed throughout these experiments.
For aesthetic enhancement, we retrain the aesthetic enhancement network using VeraRetouch-derived labels while keeping the pretrained color refinement network fixed.
For reference-guided transfer, we fit the grid LUT to the edited reference and apply the same fixed refinement network.
\Cref{tab:app_vera} shows that \method achieves the lowest color error and LPIPS and the highest PSNR and SSIM among the compared methods in both tasks.
These results demonstrate that \method adapts effectively to color adjustment labels from different retouching models while retaining a shared, fixed color refinement network, achieving higher fidelity than the compared methods in both tasks.
Target construction details are in \cref{app:experimental_settings}.

\section{Conclusion}
\label{sec:conclusion}

We presented \method, an ultralight framework for aesthetic color enhancement and reference-guided color transfer of pretrained 3DGS assets.
Our \bench benchmark enables evaluation against scene-consistent color targets across viewpoints.
Experiments show that \method achieves lower color error than competing methods, supports efficient editing, and generalizes across viewpoints, color operators, and photographic retouching priors.
These results enable practical photographic retouching of 3D assets, combining automatic aesthetic decisions with user control and geometry-preserving standard 3DGS outputs.

{
    \small
    \bibliographystyle{ieeenat_fullname}
    \bibliography{references}

@inproceedings{bychkovsky2011fivek,
  author    = {Bychkovsky, Vladimir and Paris, Sylvain and Chan, Eric and Durand, Fr{\'e}do},
  title     = {Learning Photographic Global Tonal Adjustment with a Database of Input / Output Image Pairs},
  booktitle = {CVPR},
  year      = {2011},
  pages = {97--104},
  doi = {10.1109/CVPR.2011.5995332},
}

@inproceedings{liang2021ppr10k,
  author    = {Liang, Jie and Zeng, Hui and Cui, Miaomiao and Xie, Xuansong and Zhang, Lei},
  title     = {{PPR10K}: A Large-Scale Portrait Photo Retouching Dataset with Human-Region Mask and Group-Level Consistency},
  booktitle = {CVPR},
  pages     = {653--661},
  year      = {2021}
}

@article{zhuang2025styleme3d,
  author  = {Zhuang, Cailin and Hu, Yaoqi and Zhang, Xuanyang and Cheng, Wei and Bao, Jiacheng and Liu, Shengqi and Yang, Yiying and Zeng, Xianfang and Yu, Gang and Li, Ming},
  title   = {{StyleMe3D}: Stylization with Disentangled Priors by Multiple Encoders on {3D} {Gaussians}},
  journal = {arXiv preprint arXiv:2504.15281},
  year    = {2026},
  url     = {https://arxiv.org/abs/2504.15281v2}
}

@inproceedings{xiao2025multiview,
  author    = {Xiao, Lu and Wu, Jiahao and Wang, Zhanke and Wu, Guanhua and Liu, Runling and Wang, Zhiyan and Wang, Ronggang},
  title     = {Multi-View Image Enhancement Inconsistency Decoupling Guided {3D} {Gaussian} Splatting},
  booktitle = {ICASSP},
  pages     = {1--5},
  year      = {2025},
  doi       = {10.1109/ICASSP49660.2025.10890465}
}

@inproceedings{cui2025luminancegs,
  author    = {Cui, Ziteng and Chu, Xuangeng and Harada, Tatsuya},
  title     = {{Luminance-GS}: Adapting {3D} {Gaussian} Splatting to Challenging Lighting Conditions with View-Adaptive Curve Adjustment},
  booktitle = {Proceedings of the IEEE/CVF Conference on Computer Vision and Pattern Recognition (CVPR)},
  month     = {June},
  pages     = {26472--26482},
  year      = {2025}
}

@inproceedings{zhou2025litags,
  author    = {Zhou, Han and Dong, Wei and Chen, Jun},
  title     = {{LITA-GS}: Illumination-Agnostic Novel View Synthesis via Reference-Free {3D} {Gaussian} Splatting and Physical Priors},
  booktitle = {Proceedings of the IEEE/CVF Conference on Computer Vision and Pattern Recognition (CVPR)},
  month     = {June},
  pages     = {21580--21589},
  year      = {2025}
}

@article{kerbl3Dgaussians,
  author  = {Kerbl, Bernhard and Kopanas, Georgios and Leimk{\"u}hler, Thomas and Drettakis, George},
  title   = {{3D} {Gaussian} Splatting for Real-Time Radiance Field Rendering},
  journal = {ACM Transactions on Graphics},
  volume  = {42},
  number  = {4},
  year    = {2023},
  pages = {1--14},
  doi = {10.1145/3592433},
}

@article{zeng2020learning,
  author  = {Zeng, Hui and Cai, Jianrui and Li, Lida and Cao, Zisheng and Zhang, Lei},
  title   = {Learning Image-adaptive {3D} Lookup Tables for High Performance Photo Enhancement in Real-time},
  journal = {IEEE Transactions on Pattern Analysis and Machine Intelligence},
  volume  = {44},
  number  = {4},
  pages   = {2058--2073},
  year    = {2022},
  doi = {10.1109/TPAMI.2020.3026740},
}

@inproceedings{yang2022adaint,
  author    = {Yang, Canqian and Jin, Meiguang and Jia, Xu and Xu, Yi and Chen, Ying},
  title     = {{AdaInt}: Learning Adaptive Intervals for {3D} Lookup Tables on Real-Time Image Enhancement},
  booktitle = {CVPR},
  year      = {2022},
  pages = {17522--17531},
}

@inproceedings{yang2022seplut,
  author    = {Yang, Canqian and Jin, Meiguang and Xu, Yi and Zhang, Rui and Chen, Ying and Liu, Huaida},
  title     = {{SepLUT}: Separable Image-Adaptive Lookup Tables for Real-Time Image Enhancement},
  booktitle = {ECCV},
  year      = {2022},
  pages = {201--217},
  doi = {10.1007/978-3-031-19797-0_12},
}

@inproceedings{serrano2024namedcurves,
  author    = {Serrano-Lozano, David and Herranz, Luis and Brown, Michael S. and Vazquez-Corral, Javier},
  title     = {{NamedCurves}: Learned Image Enhancement via Color Naming},
  booktitle = {Computer Vision -- ECCV 2024},
  year      = {2025},
  doi = {10.1007/978-3-031-73209-6_6},
  pages = {92--108},
}

@article{wang2024bilarf,
  author  = {Wang, Yuehao and Wang, Chaoyi and Gong, Bingchen and Xue, Tianfan},
  title   = {Bilateral Guided Radiance Field Processing},
  journal = {ACM Transactions on Graphics},
  volume  = {43},
  number  = {4},
  year    = {2024},
  doi = {10.1145/3658148},
  pages = {1--13},
}

@inproceedings{instantcolorization,
  author       = {Lieber, Daniel and Mock, Alexander and Wandel, Nils},
  title        = {Instant Colorization of {G}aussian Splats},
  booktitle    = {2026 IEEE/CVF Conference on Computer Vision and Pattern Recognition--FINDINGS Track (CVPRF)},
  year         = {2026},
  doi = {10.1109/CVPRF71909.2026.00021},
  pages = {170--180},
}

@misc{saad,
  author       = {Cao, Long and Wang, Zhongquan and Li, Jie and Chen, Yuhan and Qian, Kefei and Huang, Xiangfei and Li, Guofa},
  title        = {From Sparse and Imperfect {2D} Anchors to Consistent {3D} {Gaussian} Street Scenes: Support-Aware Appearance},
  howpublished = {arXiv:2606.26007},
  year         = {2026},
  url = {https://arxiv.org/abs/2606.26007},
}

@article{reparamgs,
  author    = {Chao, Cheng-Kang Ted and Gingold, Yotam},
  title     = {Reparametrizing {3D} {G}aussian Splatting for Real-Time Palette-based Color and Luminance Editing},
  journal   = {arXiv preprint arXiv:2609.03897},
  year      = {2026},
  url       = {https://arxiv.org/abs/2609.03897},
}

@article{virgi,
  author       = {Mazzucchelli, Alessio and Ojeda-Martin, Ivan and Rivas-Manzaneque, Fernando and Garces, Elena and Penate-Sanchez, Adrian and Moreno-Noguer, Francesc},
  title        = {{VIRGi}: View-dependent Instant Recoloring of {3D} {Gaussians} Splats},
  journal      = {IEEE Transactions on Pattern Analysis and Machine Intelligence},
  doi          = {10.1109/TPAMI.2026.3665650},
  year         = {2026}
}

@inproceedings{ling2024dl3dv,
  author    = {Ling, Lu and Sheng, Yichen and Tu, Zhi and Zhao, Wentian and Xin, Cheng and Wan, Kun and Yu, Lantao and Guo, Qianyu and Yu, Zixun and Lu, Yawen and Li, Xuanmao and Sun, Xingpeng and Ashok, Rohan and Mukherjee, Aniruddha and Kang, Hao and Kong, Xiangrui and Hua, Gang and Zhang, Tianyi and Benes, Bedrich and Bera, Aniket},
  title     = {{DL3DV-10K}: A Large-Scale Scene Dataset for Deep Learning-based {3D} Vision},
  booktitle = {CVPR},
  year      = {2024},
  pages = {22160--22169},
}

@inproceedings{barron2022mipnerf360,
  author    = {Barron, Jonathan T. and Mildenhall, Ben and Verbin, Dor and Srinivasan, Pratul P. and Hedman, Peter},
  title     = {{Mip-NeRF} 360: Unbounded Anti-Aliased Neural Radiance Fields},
  booktitle = {CVPR},
  year      = {2022},
  pages = {5470--5479},
}

@inproceedings{chen2026omni3dedit,
  author    = {Chen, Liyi and Wang, Pengfei and Zhang, Guowen and Ma, Zhiyuan and Zhang, Lei},
  title     = {{Omni-3DEdit}: Generalized Versatile {3D} Editing in One-Pass},
  booktitle = {CVPR},
  year      = {2026},
  url = {https://openaccess.thecvf.com/content/CVPR2026/papers/Chen_Omni-3DEdit_Generalized_Versatile_3D_Editing_in_One-Pass_CVPR_2026_paper.pdf},
  pages = {12640--12650},
}

@inproceedings{chen2026vip3de,
  author    = {Chen, Liyi and Li, Ruihuang and Zhang, Guowen and Wang, Pengfei and Zhang, Lei},
  title     = {Fast Multi-view Consistent {3D} Editing with Video Priors},
  booktitle = {Proceedings of the AAAI Conference on Artificial Intelligence},
  year      = {2026},
  volume = {40},
  number = {4},
  pages = {2948--2956},
  doi = {10.1609/aaai.v40i4.37286},
}

@inproceedings{mazzucchelli2024irene,
  author    = {Mazzucchelli, Alessio and Garcia-Garcia, Adrian and Garces, Elena and Rivas-Manzaneque, Fernando and Moreno-Noguer, Francesc and Penate-Sanchez, Adrian},
  title     = {{IReNe}: Instant Recoloring of Neural Radiance Fields},
  booktitle = {CVPR},
  year      = {2024},
  pages = {5937--5946},
}

@article{fpgs,
  title={{FPGS: Feed-Forward Semantic-aware Photorealistic Style Transfer of Large-Scale {Gaussian} Splatting}},
  author={Kim, GeonU and {Kim Youwang} and {Lee Hyoseok} and Oh, Tae-Hyun},
  journal={International Journal of Computer Vision},
  volume={134},
  number={5},
  year={2026},
  doi={10.1007/s11263-026-02764-9},
  publisher={Springer},
  note = {Article 214},
}

@techreport{cie2019cielab,
  author      = {{International Commission on Illumination}},
  title       = {Colorimetry --- Part 4: {CIE} 1976 {L*a*b*} Colour Space},
  institution = {ISO/CIE},
  number      = {ISO/CIE 11664-4:2019},
  year        = {2019},
  url         = {https://www.cie.co.at/publications/colorimetry-part-4-cie-1976-lab-colour-space-1}
}

@article{wang2004ssim,
  author  = {Wang, Zhou and Bovik, Alan C. and Sheikh, Hamid R. and Simoncelli, Eero P.},
  title   = {Image Quality Assessment: From Error Visibility to Structural Similarity},
  journal = {IEEE Transactions on Image Processing},
  volume  = {13},
  number  = {4},
  pages   = {600--612},
  year    = {2004},
  doi     = {10.1109/TIP.2003.819861}
}

@inproceedings{zhang2018lpips,
  author    = {Zhang, Richard and Isola, Phillip and Efros, Alexei A. and Shechtman, Eli and Wang, Oliver},
  title     = {The Unreasonable Effectiveness of Deep Features as a Perceptual Metric},
  booktitle = {Proceedings of the IEEE Conference on Computer Vision and Pattern Recognition},
  pages     = {586--595},
  year      = {2018}
}

@InProceedings{video_color_grading,
    author    = {Shin, Seunghyun and Shin, Dongmin and Shin, Jisu and Jeon, Hae-Gon and Lee, Joon-Young},
    title     = {{Video Color Grading via Look-Up Table Generation}},
    booktitle = {Proceedings of the IEEE/CVF International Conference on Computer Vision (ICCV)},
    month     = {October},
    year      = {2025},
    pages     = {19141--19152}
}

@inproceedings{veraretouch,
  series={SIGGRAPH Conference Papers '26},
  title={{VeraRetouch: A Lightweight Fully Differentiable Framework for Multi-Task Reasoning Photo Retouching}},
  url={http://dx.doi.org/10.1145/3799902.3811065},
  DOI={10.1145/3799902.3811065},
  booktitle={SIGGRAPH 2026 Conference Papers},
  publisher={ACM},
  author={Guo, Yihong and Lyu, Youwei and Tang, Jiajun and Zhou, Yizhuo and Wang, Hongliang and Chen, Jinwei and Zou, Changqing and Fan, Qingnan},
  year={2026},
  month={July},
  pages={1--12},
  collection={SIGGRAPH Conference Papers '26}
}

@inproceedings{xue2026glut,
  title={{GLUT}: {3D} {Gaussian} Lookup Table for Continuous Color Transformation},
  author={Xue, Danna and Serrano-Lozano, David and Su, Shaolin and Vazquez-Corral, Javier},
  booktitle={Advances in Neural Information Processing Systems (NeurIPS)},
  year={2026},
  url = {https://neurips.cc/virtual/2026/poster/153974},
}

@inproceedings{loshchilov2019decoupled,
  author    = {Loshchilov, Ilya and Hutter, Frank},
  title     = {Decoupled Weight Decay Regularization},
  booktitle = {International Conference on Learning Representations},
  year      = {2019},
  url       = {https://openreview.net/forum?id=Bkg6RiCqY7}
}

@inproceedings{smith2019superconvergence,
  author    = {Smith, Leslie N. and Topin, Nicholay},
  title     = {Super-convergence: very fast training of neural networks using large learning rates},
  booktitle = {Artificial Intelligence and Machine Learning for Multi-Domain Operations Applications},
  editor    = {Pham, Tien},
  volume    = {11006},
  publisher = {SPIE},
  year      = {2019},
  month     = {May},
  doi       = {10.1117/12.2520589},
  url       = {https://doi.org/10.1117/12.2520589}
}

@inproceedings{kingma2015adam,
  author = {Kingma, Diederik P. and Ba, Jimmy},
  title = {{Adam}: A Method for Stochastic Optimization},
  booktitle = {International Conference on Learning Representations},
  year = {2015},
  url = {https://arxiv.org/abs/1412.6980}
}

@article{hendrycks2016gelu,
  author = {Hendrycks, Dan and Gimpel, Kevin},
  title = {Gaussian Error Linear Units ({GELUs})},
  journal = {arXiv preprint arXiv:1606.08415},
  year = {2016},
  url = {https://arxiv.org/abs/1606.08415}
}

@inproceedings{krizhevsky2012imagenet,
  author = {Krizhevsky, Alex and Sutskever, Ilya and Hinton, Geoffrey E.},
  title = {{ImageNet} Classification with Deep Convolutional Neural Networks},
  booktitle = {Advances in Neural Information Processing Systems},
  volume = {25},
  year = {2012},
  url = {https://proceedings.neurips.cc/paper/2012/hash/c399862d3b9d6b76c8436e924a68c45b-Abstract.html}
}
}
\appendix
\crefalias{section}{appendix}
\crefalias{subsection}{appendix}
\crefalias{subsubsection}{appendix}
\twocolumn[{
    \centering
    {\Large\bfseries Appendix\par}
    \vspace{0.5em}
    {\Large\bfseries\thetitle\par}
    \vspace{1em}
}]
\section{\bench Benchmark Details}
\label{app:benchmark}

\subsection{Scene Assets and Statistics}
\label{app:benchmark_scenes}

Aesthetic color enhancement and reference-guided color transfer use the same 37 test assets in \bench: 28 scenes from DL3DV~\cite{ling2024dl3dv} and all nine scenes from Mip-NeRF 360~\cite{barron2022mipnerf360}.
The Mip-NeRF 360 subset contains four indoor scenes, \emph{room}, \emph{counter}, \emph{kitchen}, and \emph{bonsai}, and five outdoor scenes, \emph{bicycle}, \emph{flowers}, \emph{garden}, \emph{stump}, and \emph{treehill}.
DL3DV assets come from the publicly released \href{https://huggingface.co/datasets/ShapeSplats/dl3dv_10k_3dgs_new}{DL3DV 3DGS asset collection}.
We evaluate the complete assets and report supplementary breakdowns by their two scene sources.
Reconstruction-viewpoint counts in \cref{tab:benchmark_stats} describe the full set of viewpoints supplied with each asset. Evaluation viewpoint subsets are defined in \cref{app:benchmark_cameras}.

\Cref{tab:benchmark_stats} summarizes asset scale and target edit magnitude, and \cref{tab:benchmark_scene_statistics} provides the per-scene statistics.
We render DL3DV at $960\times540$ pixels and Mip-NeRF 360 at width 960 with scene-dependent heights between 622 and 640 pixels.
Focal lengths scale with image dimensions, and the principal point is the image center.
All renders use a black background.
Edit magnitude measures the original-to-target \dE averaged over the 16 off-trajectory views within each scene.
Summaries for each scene group report the mean and population standard deviation of these scene means.

\begin{table*}[t]
    \centering
    \small
    \setlength{\tabcolsep}{10pt}
    \begin{tabular}{lrrrr}
        \toprule
        Scene source & Test scenes & Gaussians (millions) & Reconstruction viewpoints & Edit magnitude (\dE) \\
        \midrule
        DL3DV & 28 & $0.95\ [0.28,\,4.39]$ & $336.5\ [302,\,421]$ & $9.66\pm2.78$ \\
        Mip-NeRF 360 & 9 & $3.64\ [1.22,\,6.13]$ & $194\ [125,\,311]$ & $13.38\pm2.04$ \\
        \bottomrule
    \end{tabular}
    \caption{\bench test-scene statistics. Gaussian and viewpoint counts report median [minimum, maximum]. Edit magnitude reports mean $\pm$ standard deviation.}
    \label{tab:benchmark_stats}
\end{table*}

\begin{table*}[!t]
    \centering
    \footnotesize
    \setlength{\tabcolsep}{6pt}
    \begin{tabular}{llrrrr}
        \toprule
        Scene source & Scene & Gaussians & Viewpoints & Render size & Edit magnitude (\dE) \\
        \midrule
        DL3DV & \textit{\texttt{971fb09235c6575e}} & 1,569,484 & 337 & $960\times540$ & 10.37 \\
        DL3DV & \textit{\texttt{97b769bc525dbc21}} & 1,249,761 & 421 & $960\times540$ & 9.10 \\
        DL3DV & \textit{\texttt{97c7586ee4b3c471}} & 621,801 & 403 & $960\times540$ & 9.03 \\
        DL3DV & \textit{\texttt{97c8843407a4fef1}} & 2,033,317 & 316 & $960\times540$ & 6.05 \\
        DL3DV & \textit{\texttt{97cb739ab732aba6}} & 1,357,464 & 317 & $960\times540$ & 15.20 \\
        DL3DV & \textit{\texttt{98ff6a357347abc2}} & 2,388,824 & 337 & $960\times540$ & 13.70 \\
        DL3DV & \textit{\texttt{9aaa19b7b5382820}} & 796,670 & 313 & $960\times540$ & 5.81 \\
        DL3DV & \textit{\texttt{9d4e6ce5dc227f7f}} & 828,309 & 388 & $960\times540$ & 10.61 \\
        DL3DV & \textit{\texttt{9df1d33910dfbee8}} & 1,039,748 & 327 & $960\times540$ & 8.40 \\
        DL3DV & \textit{\texttt{9e7a4d73f72909ea}} & 1,051,898 & 318 & $960\times540$ & 11.64 \\
        DL3DV & \textit{\texttt{9f36659117da4b97}} & 870,244 & 416 & $960\times540$ & 10.25 \\
        DL3DV & \textit{\texttt{a141d53dc0f0cf2c}} & 1,171,531 & 332 & $960\times540$ & 6.52 \\
        DL3DV & \textit{\texttt{a1be174f991bd9e4}} & 467,932 & 344 & $960\times540$ & 5.88 \\
        DL3DV & \textit{\texttt{a2114ac1ac723af7}} & 723,333 & 302 & $960\times540$ & 5.73 \\
        DL3DV & \textit{\texttt{a2ff43318ac51b73}} & 1,685,162 & 324 & $960\times540$ & 12.70 \\
        DL3DV & \textit{\texttt{a48925bc8ab540cb}} & 632,712 & 335 & $960\times540$ & 10.57 \\
        DL3DV & \textit{\texttt{a4e3a36e5afc4bf8}} & 379,267 & 337 & $960\times540$ & 10.98 \\
        DL3DV & \textit{\texttt{a602e1418d1697b2}} & 1,679,032 & 307 & $960\times540$ & 5.94 \\
        DL3DV & \textit{\texttt{a6c3d1408437e10b}} & 600,862 & 340 & $960\times540$ & 14.68 \\
        DL3DV & \textit{\texttt{a750134923b8a7de}} & 424,820 & 404 & $960\times540$ & 14.28 \\
        DL3DV & \textit{\texttt{a79a53e3dfdf60cc}} & 341,855 & 336 & $960\times540$ & 10.71 \\
        DL3DV & \textit{\texttt{ab23945f82e7ede2}} & 565,490 & 314 & $960\times540$ & 5.86 \\
        DL3DV & \textit{\texttt{ad2a4e43fcea11a0}} & 1,103,898 & 341 & $960\times540$ & 11.15 \\
        DL3DV & \textit{\texttt{ad76315256f4d737}} & 546,417 & 404 & $960\times540$ & 8.74 \\
        DL3DV & \textit{\texttt{ae1c989bc6b416e5}} & 1,150,280 & 362 & $960\times540$ & 9.60 \\
        DL3DV & \textit{\texttt{af10d40591e57fab}} & 3,178,076 & 382 & $960\times540$ & 8.90 \\
        DL3DV & \textit{\texttt{af240b51fb1c3b7d}} & 279,270 & 320 & $960\times540$ & 8.21 \\
        DL3DV & \textit{\texttt{af645f082f54c231}} & 4,391,780 & 329 & $960\times540$ & 9.79 \\
        \midrule
        Mip-NeRF 360 & \textit{bicycle} & 6,131,954 & 194 & $960\times638$ & 15.15 \\
        Mip-NeRF 360 & \textit{bonsai} & 1,244,819 & 292 & $960\times640$ & 14.11 \\
        Mip-NeRF 360 & \textit{counter} & 1,222,956 & 240 & $960\times640$ & 12.74 \\
        Mip-NeRF 360 & \textit{flowers} & 3,636,448 & 173 & $960\times633$ & 11.39 \\
        Mip-NeRF 360 & \textit{garden} & 5,834,784 & 185 & $960\times622$ & 13.85 \\
        Mip-NeRF 360 & \textit{kitchen} & 1,852,335 & 279 & $960\times640$ & 9.69 \\
        Mip-NeRF 360 & \textit{room} & 1,593,376 & 311 & $960\times640$ & 15.16 \\
        Mip-NeRF 360 & \textit{stump} & 4,961,797 & 125 & $960\times636$ & 16.55 \\
        Mip-NeRF 360 & \textit{treehill} & 3,783,761 & 141 & $960\times630$ & 11.78 \\
        \bottomrule
    \end{tabular}
    \caption{Per-scene statistics for all 37 assets in \bench. DL3DV scene identifiers are abbreviated to their first 16 characters. Viewpoint counts cover the full reconstruction set. Edit magnitude is the mean original-to-target \dE over the 16 off-trajectory views.}
    \label{tab:benchmark_scene_statistics}
\end{table*}

\subsection{Target Construction}
\label{app:benchmark_targets}

We use pretrained NamedCurves MIT5K-DPE~\cite{serrano2024namedcurves} to obtain the benchmark's target colors and tones.
Offline target construction jointly fits one scene-level color operator to the 20 input and 16 held-out reconstruction views of each scene.
Source renders and enhancer outputs are clipped and quantized using:
\begin{equation}
    Q_8(x)=\frac{\operatorname{round}(255\Pi(x))}{255},
    \label{eq:benchmark_quantization}
\end{equation}
where $\Pi$ clips RGB values to $[0,1]$.
For original render $I$, the paired images are $Q_8(I)$ and $Q_8(E(Q_8(I)))$, where $E$ denotes the pretrained image enhancer.
We sample 15,000 pixel locations uniformly without replacement from each image, including background pixels.
Sampling uses a fixed seed for each scene, yielding 540,000 color pairs from its 36 views.

The benchmark targets represent scene-level color and tone adjustments through a shared RGB-to-RGB mapping across viewpoints.
Each scene LUT uses a uniform $17\times17\times17$ RGB grid with $D=17^3=4,913$ LUT nodes and spacing $1/16$.
We fit LUT-node displacements from the identity mapping using the ridge objective in \cref{app:operator_fitting}, setting $\lambda_{\mathrm{LUT}}=10^{-2}\operatorname{tr}(A^\top A)/D$ and solving the normal equations directly.
The rows of $A$ contain the trilinear interpolation weights of the sampled input colors.
The fitted operator interpolates the displacements, adds them to the input RGB, and clips the result to $[0,1]$.
LUT fitting and application operate directly on rendered RGB values.
Evaluation uses floating-point targets $I_s^\star(v)=T_s(\Pi(R(\mathcal G_s,v)))$, and the edited reference image is $Q_8(I_s^\star(v_0))$.
The scene LUT and target images are reserved for evaluation. Transfer methods receive the edited reference at $v_0$.

\subsection{Viewpoint Splits}
\label{app:benchmark_cameras}

Each scene has 20 input and 16 held-out reconstruction viewpoints.
DL3DV uses fixed viewpoint lists.
For Mip-NeRF 360, a random permutation with seed zero assigns the first 20 reconstruction viewpoints to the input set and the next 16 to the held-out set.
These sets are disjoint, and $v_0$ is the first input viewpoint.
All viewpoint lists are fixed before evaluation.
The benchmark protocol records the scene identifiers, input and held-out view indices, reference view, and complete off-trajectory camera parameters.
Evaluation reads the stored camera parameters directly to reproduce the same viewpoints.

We generate one off-trajectory viewpoint from each held-out viewpoint.
Let $t_j$ denote the reconstruction camera centers and $\bar t$ their mean. The radius $r_s=1.1\max_j\|t_j-\bar t\|_2$ defines a sphere centered at $\bar t$ that encloses these centers.
We sample a rotation angle uniformly from $[10^\circ,20^\circ]$ around an axis whose azimuth is uniform in $[0,2\pi)$ in the camera's local image plane.
For translation, we draw a three-dimensional standard normal vector, multiply its local depth component by $0.3$, normalize it, and transform it to world coordinates.
The translation length is $0.075r_s$.
Perturbations use fixed seeds for each scene and held-out viewpoint.
Intrinsics and image dimensions remain fixed.
The resulting 16 off-trajectory poses are reserved for evaluating the delivered asset.
Together, the 20 input, 16 held-out, and 16 off-trajectory viewpoints form three evaluation groups shared by both tasks.

\subsection{Metrics and Aggregation}
\label{app:benchmark_metrics}

Let $\hat I_s(v)$ denote the edited asset's render and $P_v$ all pixels in view $v$, including the background.
We clip RGB images to $[0,1]$ and convert sRGB to CIELAB under the D65 white point~\cite{cie2019cielab}.
The conversion uses the standard sRGB inverse transfer function and white point $(X_n,Y_n,Z_n)=(0.95047,1,1.08883)$.
The per-view color error is:
\begin{equation}
    d_s(v)=\frac{1}{|P_v|}\sum_{p\in P_v}
    \left\|\operatorname{Lab}(\Pi(\hat I_s(v))_p)
    -\operatorname{Lab}(I_s^\star(v)_p)\right\|_2.
    \label{eq:benchmark_color_error}
\end{equation}
For an evaluation viewpoint set $\mathcal V_s$, the scene score is $|\mathcal V_s|^{-1}\sum_{v\in\mathcal V_s}d_s(v)$.
The overall \bench score averages all 37 scene scores equally.
Breakdowns by scene source use the same equal-scene weighting within each group.
The primary score uses the 16 off-trajectory viewpoints $\mathcal V_s^{\mathrm{off}}$.
The view-generalization gap is:
\begin{equation}
    g_s=\frac{1}{16}\sum_{v\in\mathcal V_s^{\mathrm{off}}}d_s(v)-d_s(v_0).
    \label{eq:benchmark_generalization_gap}
\end{equation}
Both tasks use the same reference viewpoint $v_0$.
We aggregate $g_s$ with the same equal-scene weighting as color error.

\section{Method Implementation Details}
\label{app:implementation}

\subsection{Operator Fitting}
\label{app:operator_fitting}

The predicted enhancement LUT and the reference-fitted transfer LUT use a uniform $17^3$ RGB grid, giving $D=4{,}913$ LUT nodes storing RGB displacements and grid spacing $1/16$.
Queries use trilinear interpolation and output clipping as defined in \cref{eq:operator}.
For reference-guided color transfer, the original rendering is clipped to $[0,1]$ and paired with the edited reference pixels.
The fitted displacements minimize:
\begin{equation}
    \widehat\Theta=\arg\min_\Theta
    \|A\Theta-(Y-X)\|_F^2+\lambda_{\mathrm{LUT}}\|\Theta\|_F^2,
    \label{eq:reference_fit}
\end{equation}
where $X$ and $Y$ contain paired RGB samples and the rows of $A$ are the interpolation weights of $X$.
The identity-centered penalty anchors colors sparsely represented in the reference.
Images with more than 60,000 pixels are sampled at 60,000 evenly spaced flattened indices, rounded down to integers.
The ridge coefficient in \cref{eq:reference_fit} is $\lambda_{\mathrm{LUT}}=10^{-2}\max(\operatorname{tr}(A^\top A)/D,10^{-9})$.
The normal equations use diagonal preconditioner $\operatorname{diag}(A^\top A)+\lambda_{\mathrm{LUT}}$ and conjugate gradients initialized at zero.
The solve stops after 80 iterations or when the joint RGB residual norm falls below $10^{-7}$.
Training labels follow the multi-view target construction in \cref{app:benchmark_targets}.

\subsection{Projection and Network Details}
\label{app:projection}

\noindent{\textbf{Gaussian Selection and Projection.}}
Using the associated camera parameters $\mathcal C$, the aesthetic enhancement network selects up to 262,144 Gaussians with the largest summed projected footprints over the 20 input viewpoints.
Let $r_i=\sqrt{s_i^{(1)}s_i^{(2)}}$ use Gaussian $i$'s two largest scales, $o_i$ denote opacity, and $d_{iv}$ denote camera-coordinate depth at viewpoint $v$.
For focal length $f_{xv}$ and grid dimensions $W_v,H_v$, the footprint is:
\begin{equation}
F_{iv}=\begin{cases}
\min\!\left(o_i(f_{xv}r_i/d_{iv})^2,W_vH_v\right),&\text{visible},\\
0,&\text{otherwise}.
\end{cases}
\label{eq:app_footprint}
\end{equation}
A center is visible when its depth exceeds $10^{-3}$ and its projection lies inside the image.
Projection grids have width 80 and preserve aspect ratio.
Camera intrinsics, including the focal length used in the footprint, are scaled to this grid resolution before projection.
Flooring projected coordinates assigns cell $g_v(i)$.
For visible centers, the aggregation weight is:
\begin{equation}
\omega_{iv}=o_i\exp\!\left[-\frac{\log d_{iv}-\log d^{\min}_{v,g_v(i)}}{0.05}\right],
\label{eq:app_projection_weight}
\end{equation}
where $d^{\min}_{vg}$ is the nearest center depth in cell $g$.
The aesthetic enhancement network computes these weights within its selected Gaussian set.

\noindent{\textbf{Aesthetic Enhancement Network.}}
A $7\!\to\!64\!\to\!32$ MLP encodes view-dependent SH RGB, opacity, and three relative log scales.
The log scales subtract their median over the selected set, are sorted in descending order, clipped to $[-6,6]$, and divided by three.
SH colors use $Q_8$ quantization. Training features store opacity and scales in half precision.
Each cell contains the weighted feature mean, with $10^{-6}$ added to its denominator, and density channel $\log(1+\sum_i\omega_{iv})$.
Three convolutions map channels as $33\!\to\!64\!\to\!64\!\to\!64$, with kernels $3,3,1$, strides $1,2,1$, and padding $1,1,0$.
Spatial mean and maximum pooling provide 128 features.
The same grid aggregates the selected Gaussians' quantized view-dependent SH colors using $\omega_{iv}$, with a $10^{-12}$ floor on the summed-weight denominator.
Occupied cells contribute equally to five quantiles: 2nd, 25th, 50th, 75th, and 98th percentiles.
These quantiles of R, G, B, and luma $0.299R+0.587G+0.114B$ provide 20 additional features.
A $148\!\to\!128\!\to\!128$ MLP encodes the features at each viewpoint, followed by averaging across viewpoints and a $128\!\to\!128\!\to\!128\!\to\!32$ coefficient head.
MLP hidden layers and the first two convolutions use GELU~\cite{hendrycks2016gelu}.
The PCA basis has 32 columns, formed from the leading right singular vectors of the centered training-LUT matrix.
For the 189 training LUTs, each coefficient standard deviation is its singular value divided by $\sqrt{188}$.
The head's final weight and bias start at zero, giving the training-LUT mean as the initial prediction.

\noindent{\textbf{Blended Color Context.}}
Blended color context construction uses all Gaussians and the projection weights in \cref{eq:app_projection_weight}.
Let $\bar c_j(v)$ be the view-dependent SH color clamped below at zero and $W_i=\sum_{v\in\mathcal C}\omega_{iv}$.
Grid colors and blended color context are:
\begin{equation}
\begin{aligned}
q_{vg}&=\frac{\sum_{j:g_v(j)=g}\omega_{jv}\bar c_j(v)}{\max(\sum_{j:g_v(j)=g}\omega_{jv},10^{-12})},\\
m_i&=\begin{cases}
\Pi\!\left(\dfrac{\sum_{v\in\mathcal C}\omega_{iv}q_{v,g_v(i)}}{W_i}\right),&W_i>10^{-6},\\
u_i,&W_i\leq10^{-6}.
\end{cases}
\end{aligned}
\label{eq:app_context}
\end{equation}

\subsection{Color Update Implementation}
\label{app:writeback}

The blend-aware color refinement network in \cref{sec:method_writer} predicts the DC correction $d_{\theta,i}$ through two residual-propagation layers.
Its graph connects each Gaussian to eight Euclidean nearest neighbors.
Its 16-dimensional node input contains $\tilde c_i$, $m_i$, their difference, $a_i$, $o_i$, three relative log scales, $\log(1+\|h_{i,\ell>0}\|_F)$, and $\log(1+W_i)/3$.
Relative log scales use the aesthetic enhancement network's normalization, computed over the full asset.
LayerNorm and a $16\!\to\!64\!\to\!64$ MLP encode each node.
A residual context MLP with LayerNorm and widths $128\!\to\!64\!\to\!64$ combines its embedding with the mean neighbor embedding.
Both propagation layers reuse the resulting node representations, and each self head takes the corresponding 64-dimensional node embedding.

For centers $p_i,p_j$, the geometry channels use:
\begin{equation}
\begin{aligned}
\rho_{ij}&=\frac{p_j-p_i}{\max(r_i,10^{-6})},\\
g_{ij}&=-\frac{\|p_j-p_i\|_2^2}{\max(r_i^2+r_j^2,10^{-10})},\\
\xi_{ij}&=\frac15[\operatorname{clip}(\rho_{ij},-20,20),
\operatorname{clip}(g_{ij},-20,0)].
\end{aligned}
\label{eq:app_edge_geometry}
\end{equation}
Each edge head receives two 64-dimensional embeddings, $\xi_{ij}$, and an additional copy of its clipped distance channel $\operatorname{clip}(g_{ij},-20,0)/5$, totaling 133 inputs.
Self and edge heads predict the scalar coefficients that combine operator residuals and blending discrepancy features in \cref{eq:writeback}.
Self heads output four coefficients in the first layer and five in the second.
Edge heads output five and six, respectively, including the coefficient for the edge blending discrepancy feature.
All heads have 64 hidden units with GELU and zero-initialized output layers.

Let $b_i=b_T(u_i,m_i)$, $b_{ij}=b_T(u_i,u_j)$, and $\mathcal R_i=(r_i^{\mathrm{dir}},r_i^{\mathrm{ctx}},r_i^{\mathrm{jac}},b_i)$ denote the ordered residual tuple.
The vectors indexed by $k$ in \cref{eq:writeback} are:
\[
\begin{aligned}
(x_{ii,k}^{(1)})_k&=\mathcal R_i,\\
(x_{ij,k}^{(1)})_k&=(\mathcal R_j,b_{ij}),\\
(x_{ii,k}^{(2)})_k&=(\mathcal R_i,d_i^{(1)}),\\
(x_{ij,k}^{(2)})_k&=(\mathcal R_j,d_j^{(1)},b_{ij}),
\end{aligned}
\]
where $j\in\mathcal N(i)$ and $d_i^{(1)}$ is the first propagation layer's correction.
Parentheses concatenate the listed vectors in order.
Neighbor terms use the residual vectors at node $j$ and are summed over the eight neighbors without normalization.
The final correction sums the outputs of both layers.

The analytic color update uses $\delta=0.5$ and a boundary-adjusted central-difference Jacobian with step $\epsilon=1/34$.
For $x_k^\pm=\Pi(x\pm\epsilon e_k)$, its $k$th column is $(T(x_k^+)-T(x_k^-))/\max([x_k^+-x_k^-]_k,10^{-6})$.
The blend-aware color refinement network uses $\tilde c_i=\operatorname{clip}(c_i,-0.5,1.5)$ and $\widetilde{J_T(m_i)-I}=\operatorname{clip}(J_T(m_i)-I,-5,5)$ elementwise.
The analytic color update uses the original DC color and the unclipped Jacobian.
The input DC color is $c_i=C_0h_{i0}+0.5$, where $h_{i0}$ is the zeroth-order SH coefficient and $C_0=1/\sqrt{4\pi}$.
After the color update, the DC SH coefficient is $h'_{i0}=(c_i'-0.5)/C_0$.

\subsection{Solver Demonstrations}
\label{app:solver_demonstrations}

A solver demonstration provides target DC corrections for training the blend-aware color refinement network with \cref{eq:distill_loss}.
Each demonstration pairs a scene with a color operator sampled using the LUT interpolation and displacement scaling specified in \cref{app:training}.
The visibility-weighted solver of Instant Colorization~\cite{instantcolorization} starts from the SH coefficients $H^{\mathrm{cf}}$ produced by the analytic color update and uses all 20 input viewpoints.
Let $H_i\in\mathbb R^{M\times3}$ contain Gaussian $i$'s $M$ SH coefficients, and let $R(H,v)$ render the fixed geometry and opacity with coefficients $H$.
The target is $Y_v=T(\Pi(R(\mathcal G,v)))$, and the data loss is:
\begin{equation}
\mathcal L_{\mathrm{demo}}(H)
=\frac12\sum_{v\in\mathcal C}\|R(H,v)-Y_v\|_F^2.
\label{eq:app_demo_loss}
\end{equation}
This loss sums errors over views, pixels, and RGB channels.
For compositing weights $w_{ivp}$, define $a_{iv}=\sum_p w_{ivp}$ and total visibility $\nu_i=\sum_v a_{iv}$.
Let $y_{iv}\in\mathbb R^M$ be the SH basis evaluated in Gaussian $i$'s viewing direction.
For $\nu_i>10^{-6}$, the SH regularization matrix is $\Gamma_i=\nu_i\operatorname{diag}(\lambda_\ell)$, where $\lambda_0=10^{-5}$ and $\lambda_\ell=100$ for $\ell>0$.
Starting from $H^{(0)}=H^{\mathrm{cf}}$, the solver performs:
\begin{equation}
\begin{aligned}
P_i&=\sum_{v\in\mathcal C}a_{iv}y_{iv}y_{iv}^{\top}+\Gamma_i,\\
b_i^{(t)}&=-\nabla_{H_i}\mathcal L_{\mathrm{demo}}(H^{(t)})\\
&\quad-\Gamma_i(H_i^{(t)}-H_i^{\mathrm{cf}}),\\
H_i^{(t+1)}&=H_i^{(t)}+P_i^{-1}b_i^{(t)}.
\end{aligned}
\label{eq:app_demo_solver}
\end{equation}
We factor each $P_i$ once with Cholesky decomposition and reuse its factors for ten updates.
The final demonstration uses $H_i^{\mathrm{cf}}$ for Gaussians with $\nu_i\leq10^{-6}$.
The supervised DC correction is $d_i^\star=C_0(H_{i0}^{(10)}-H_{i0}^{\mathrm{cf}})$ and is stored in half precision.
Distillation uses $\eta_i=\nu_i/(\nu_i+\tilde\nu)$, where $\tilde\nu$ is the median visibility over Gaussians with $\nu_i>10^{-6}$.
The distillation loss sums squared RGB errors per Gaussian and normalizes their visibility-weighted sum by $\sum_i\eta_i$.

\section{Experimental Settings}
\label{app:experimental_settings}

\subsection{\method Training Settings}
\label{app:training}

\noindent{\textbf{Training Data.}}
The aesthetic enhancement network uses 189 DL3DV~\cite{ling2024dl3dv} training scenes and 23 validation scenes.
Of the 189 training scenes, 145 supply the LUTs used to train the blend-aware color refinement network.
Training, validation, and test scenes are disjoint.
The blend-aware color refinement network uses 16 complete assets from this 145-scene subset and six validation assets, selected with seed zero.
Training and validation assets contain at most 800,000 and 1,000,000 Gaussians, respectively.
Each selected asset contributes all its Gaussians.

The refinement operator bank contains 580 LUTs from these 145 scenes and four pretrained enhancers: the MIT5K-DPE, PPR10K-C, and MIT5K-UEGAN variants of NamedCurves~\cite{serrano2024namedcurves}, and the sRGB model of image-adaptive 3D LUT~\cite{zeng2020learning}.
Each scene-level LUT is fitted to 36 enhanced views using \cref{app:benchmark_targets}.
The aesthetic enhancement network learns the MIT5K-DPE targets.

\noindent{\textbf{Enhancement Supervision.}}
The color bank $\mathcal X_s$ in \cref{eq:enhancement_loss} contains 131,072 quantized colors per scene, sampled from pairs of Gaussians and viewpoints with replacement and probabilities proportional to $F_{iv}$.
Pairs are drawn from the selected Gaussian set and the 20 input viewpoints. Each entry stores the view-dependent SH color of its sampled pair, clipped to $[0,1]$ and quantized with $Q_8$.
The color bank remains fixed throughout training.
Each enhancement step samples four scenes with replacement and four to eight input viewpoints per scene.
For each scene, the viewpoint count is sampled uniformly, and a random permutation selects the viewpoints.
Each scene contributes 65,536 bank entries sampled with replacement per step.
The enhancement loss averages absolute error over colors and RGB channels, including the $1/3$ factor in the main text.
Inference averages the representations from all 20 input viewpoints.

\noindent{\textbf{Operator Sampling and Refinement Training.}}
During rendering fine-tuning, we sample the identity color operator with probability $0.1$.
Otherwise, we uniformly sample a bank LUT and interpolate it with another uniformly sampled bank LUT with probability $0.5$, using a coefficient sampled from $U[0,1]$.
The resulting displacements are scaled by $U[0.5,1.5]$.
Solver demonstrations use non-identity operators drawn with the same interpolation and scaling procedure and a fixed seed per scene.
Each training asset supplies 24 non-identity demonstrations, giving 384 scene-operator pairs, and each validation asset supplies eight.
The solver and visibility weights for distillation are specified in \cref{app:solver_demonstrations}.

Both refinement stages uniformly sample a scene every four updates and reuse it for those updates.
Distillation samples one solver demonstration uniformly at each update.
Rendering fine-tuning draws a new operator using the sampler above.
Rendering fine-tuning uses one scene, one operator, and two views per update, rendered at a width of 480 pixels.
Each view uses an input camera pose with probability $0.3$, or a perturbed pose otherwise.
Perturbations use the local-axis and translation-direction construction in \cref{app:benchmark_cameras}, with training ranges $U[0^\circ,20^\circ]$ for rotation and $U[0,0.075r_s]$ for translation length.
The image loss averages over sampled views, pixels, and RGB channels.
The correction penalty uses $\lambda_{\mathrm{corr}}=10^{-3}$ and averages squared corrections over Gaussians and RGB.

\noindent{\textbf{Optimization and Model Selection.}}
All stages use AdamW~\cite{loshchilov2019decoupled}, weight decay $10^{-4}$, OneCycle~\cite{smith2019superconvergence} scheduling with 5\% warm-up, and gradient-norm clipping at one.
Network initialization and training-time sampling use seed zero.
Aesthetic enhancement, solver distillation, and rendering fine-tuning use 20,000, 16,000, and 12,000 updates, with peak learning rates of $3\times10^{-4}$, $10^{-3}$, and $3\times10^{-4}$, respectively.
Each stage starts a new optimizer and schedule.
We select the aesthetic enhancement network by mean validation image-space \dE after 5,000, 10,000, and 20,000 training updates.
Validation applies the predicted and target LUTs to 20 input views per scene at width 240 and selects the model trained for 20,000 updates.
Our implementation uses Python 3.10, PyTorch 2.11.0 with CUDA 12.8, and gsplat 1.5.3.

\subsection{Inputs and Output Representations}
\label{app:baseline_interfaces}
\label{app:baselines}

All asset comparisons start from the same source assets and follow the same benchmark viewpoint protocol.
In aesthetic enhancement, image-conditioned baselines obtain their targets by applying NamedCurves MIT5K-DPE~\cite{serrano2024namedcurves} to input views.
Instant Colorization, BilaRF, ReparamGS, FPGS, VIRGi, ViP3DE, and Omni-3DEdit receive one enhanced view, while SAAD receives sixteen.
FPGS and ViP3DE also read twenty original input-view renders.
ReparamGS and VIRGi use original renders for scene preparation, as detailed below.
These preparation views are separate from the enhanced target images used to supervise editing.
VCG is evaluated as an alternative LUT source in \cref{app:alternative_operators}.
In reference-guided color transfer, each method receives the same edited image at $v_0$.
The reference image supplies the intended color appearance for the entire scene.

We evaluate each method's native output.
Standard 3DGS assets use the common 3DGS renderer.
BilaRF preserves its bilateral grid, ReparamGS preserves palette-weight SH and tone controls, and VIRGi preserves its neural color representation.
Omni-3DEdit produces edited images.

\subsection{Asset Editing Baselines}
\label{app:asset_baselines}

\noindent{\textbf{Instant Colorization.}}
Instant Colorization~\cite{instantcolorization} uses the authors' color-statistics collection, visibility normalization, linear solve, and CUDA rasterizer.
We provide the input Gaussian geometry and registered camera parameters, then perform the initial solve and fifty refinements from one edited view.
The SH regularization weights are $10^{-5}$ for DC and $10^{-2}$ for each higher order.
Each per-Gaussian system uses a direct linear solve.
The output is a standard 3DGS asset.

\noindent{\textbf{BilaRF.}}
We couple the authors' CP-factorized 4D bilateral grid~\cite{wang2024bilarf} to the input 3DGS asset to adjust colors using 3D position and learned grayscale guidance.
The grid uses spatial resolution $16^3$, eight bins along the guidance dimension, factorization rank five, and a spatial coordinate bound of one.
Its learned grayscale guidance network has width eight and depth two.
Optimization uses Adam~\cite{kingma2015adam} for 2,500 updates, with learning rate decaying exponentially from $10^{-2}$ to $10^{-3}$, betas $(0.9,0.99)$, and $\epsilon=10^{-15}$.
The objective combines mean Charbonnier error $\sqrt{e^2+10^{-6}}$ and the authors' total-variation term with unit weight.
Positions are normalized by the center and radius of the reconstruction camera distribution, and full-frame supervision replaces random-ray batches.

\noindent{\textbf{SAAD.}}
Our implementation of SAAD~\cite{saad} keeps geometry and opacity fixed and learns scene-specific appearance predictors for illumination, tone, and detail.
Adam runs for 5,000 updates at learning rate $10^{-3}$.
The predictors use two hidden ReLU layers, widths 64 for illumination and tone and 32 for detail, four Fourier bands, and degree-three SH.
Gaussian filters use kernel size and standard deviation $(21,5)$ and $(7,1.5)$.
Aesthetic enhancement supplies sixteen evenly spaced enhanced views from the input set, while reference-guided color transfer supplies one edited supervision view.
For numerical settings unspecified by the paper, we activate the illumination and detail branches after steps 1,000 and 3,000, respectively.
The Gaussian-level auxiliary loss has weight one before the first transition and $0.1$ afterwards. The quantile threshold for selecting detail-supervision pixels decreases from $0.95$ to $0.70$ after the second transition.
The temperature controlling residual agreement across supervision views is $0.01$, and DC-magnitude, direction, and higher-order-SH penalties each have weight $10^{-4}$.
Tone statistics use global and $4\times4$ patch means and population standard deviations in BT.601 YUV.
The loss sums over pixels and supervision views, accumulating gradients from all supervision views before each update.

\noindent{\textbf{ReparamGS.}}
The authors' palette decomposition and constraint solver~\cite{reparamgs} prepare the asset for ten epochs using original renders from the reconstruction viewpoints, excluding the sixteen held-out viewpoints.
Seed 42 initializes four evenly spaced chromatic palette colors, black and white, and degree-three palette-weight SH.
Adam learning rates are $0.005$ for DC weights, $0.0005$ for higher orders, and $0.01$ for palette parameters.
The decomposition combines barycentric MSE, positivity with weight $0.5$, compactness increasing from $0.001$ to $0.005$, and unit-weight hue separation with minimum gap $0.3$.
Editing uses 100 reference pixels on a fixed $10\times10$ grid.
The constraint solver adjusts palette colors and tone curves using damped block coordinate descent (dBCD) with iteratively reweighted least squares (IRLS).
It performs up to eight outer rounds with damping $0.7$, 100 curve samples, constraint weight $w_{\mathrm{eq}}=1{,}000$, palette-shift weight $w_{\mathrm{sp}}=0.1$ within the sparsity term, $\rho=100$, and eight inner iterations with $\epsilon=0.001$.
The delivered representation preserves the palette and tone controls.

\noindent{\textbf{FPGS.}}
FPGS~\cite{fpgs} follows the authors' preprocessing, feature-autoencoder training, Gaussian reconstruction, and style-transfer procedure.
Twenty input-view renders provide the source images, and the input Gaussian centers and DC colors initialize the point cloud.
DINO features use resolution 448, overlapping $113\times113$ grids, horizontal-flip averaging, and guided filtering.
The autoencoder trains for five epochs with Adam at $0.001$.
Scene reconstruction uses 30,000 updates with RGB $L_1$ and DSSIM, adding semantic $L_1$ with weight $0.1$ after step 25,000.
Native densification and opacity resets are preserved.
Style transfer uses two passes, temperature 100, $\epsilon=0.01$, and blending $0.3$.
We evaluate the resulting native 3DGS asset.

\noindent{\textbf{VIRGi.}}
Our implementation of VIRGi~\cite{virgi} uses three-layer diffuse and specular networks of width 64, unidirectional hidden-feature connections, and a sigmoid on their summed output.
The hash grid has sixteen levels with two features per level, base resolution 16, scale 1.447, and hash-table size $2^{19}$.
Directions use nine degree-two SH coefficients.
Preparation uses 30,000 Adam updates at learning rate $0.01$, with 100 warm-up steps and learning-rate multipliers $0.33$ at steps 5,000, 15,000, and 25,000.
Each batch contains 32 groups of image patches, each grouping five $16\times16$ patches that observe the same Gaussian from different viewpoints.
The specular penalty decreases from $0.25$ to $0.05$.
Editing updates a copy of the final diffuse layer and a soft-segmentation network for 200 steps at $0.01$.
The segmentation network predicts weights that blend the original and edited diffuse colors.
Segmentation and editing settings follow IReNe~\cite{mazzucchelli2024irene}.
Original-asset renders supervise preparation, and the native neural representation is evaluated with the source geometry and opacity.

\subsection{Image Propagation Baselines}
\label{app:propagation_baselines}

\noindent{\textbf{Omni-3DEdit.}}
Omni-3DEdit~\cite{chen2026omni3dedit} uses the authors' appearance model trained for 4,000 updates and inference seed 23.
For both tasks, the image-propagation pipeline receives the original/edited reference pair and sixteen off-trajectory source renders, processed in two nine-view batches with repeated views filling the final batch.
VGGT-1B estimates poses from these images.

\noindent{\textbf{ViP3DE.}}
ViP3DE~\cite{chen2026vip3de} uses the authors' custom-edit configuration.
Twenty input-view renders are ordered along the capture trajectory with the reference first.
For reference-guided color transfer, the supplied edited reference replaces the first-frame InstructPix2Pix edit.
For aesthetic enhancement, this interface receives the output of NamedCurves on the first input view.
The remaining pipeline is shared across tasks.
SVD inversion and propagation use 25 steps, inversion strength $0.15$, and internal resolution $576\times1024$.
Asset fitting runs 600 Adam updates at $0.001$, with betas $(0.9,0.99)$, $\epsilon=10^{-15}$, seed zero, and weights ten for both $L_1$ and perceptual losses.
The native pruning and densification procedure is preserved.
The input interface preserves the benchmark image dimensions and field of view, and evaluation uses the delivered 3DGS asset.

\subsection{LUT Configurations and Retouching Priors}
\label{app:alternative_operators}

\noindent{\textbf{Alternative LUT Representation: GLUT.}}
To evaluate a continuous LUT representation for reference-guided color transfer, we replace the default grid LUT with the authors' single-LUT representation of GLUT~\cite{xue2026glut}, fitted to the same original/edited reference pair.
It uses 32 color-space Gaussians and 716 trainable parameters, with residual connection and normalized Gaussian blending.
Adam trains for twenty epochs at learning rate $0.001$, batch size 4,096, and seed 52, with cosine scheduling at every update and gradient clipping at one.
The objective combines RGB $L_1$, target-chroma-weighted hue loss with weight ten, and opacity entropy with weight $0.001$.
Following the official GLUT fitting procedure, we progressively emphasize color pairs with larger prediction errors and reuse previously selected hard samples during optimization.
GLUT fits all reference pixels, whereas the grid LUT uses the sampled ridge fit in \cref{app:operator_fitting}.
Both operators use the same blended color context, analytic color update, and trained blend-aware color refinement network.
\Cref{tab:app_glut_cross} compares the two LUT representations in operator accuracy and edited-asset fidelity.

\noindent{\textbf{Video Color Grading.}}
Video Color Grading (VCG)~\cite{video_color_grading} provides an alternative source of scene-level LUTs using the authors' pretrained GS-Extractor and L-Diffuser.
It processes twenty original input-view frames and one edited reference with 25 DDIM steps and seed 48.
The edited reference corresponds to the first input viewpoint.
Its native $16^3$ LUT is sampled on a $17^3$ grid and combined with the analytic color update for asset evaluation.
\Cref{tab:operator_replacement} compares this LUT source with the default grid-LUT configurations in both tasks.

\noindent{\textbf{VeraRetouch.}}
To evaluate generalization to a new photographic retouching prior, we construct aesthetic enhancement and reference-guided color transfer targets from VeraRetouch~\cite{veraretouch} in automatic mode with temperature zero.
The blend-aware color refinement network retains its weights learned from the four enhancers in \cref{app:training}.
For aesthetic enhancement, we retrain the aesthetic enhancement network on VeraRetouch-derived scene-level LUTs.
Scene-level targets and edited references follow the construction in \cref{app:benchmark}.
The enhanced images have width 960.
For reference-guided color transfer, the analytic and refined \method variants use the same reference-fitted grid LUT.
All evaluated pipelines use the same scene targets, source assets, and sixteen off-trajectory viewpoints on all 37 scenes in \bench.
The reference-guided color transfer pipelines receive the same edited reference for each scene.
\Cref{tab:app_vera} compares aesthetic enhancement and reference-guided color transfer under these retouching targets.

\subsection{Evaluation Metrics and Statistical Analysis}
\label{app:aux_metrics}

PSNR uses mean squared error over all sRGB pixels and channels, with unit peak intensity and MSE floored at $10^{-12}$.
SSIM~\cite{wang2004ssim} uses an $11\times11$ Gaussian window with standard deviation $1.5$, $C_1=0.01^2$, and $C_2=0.03^2$.
Convolution uses valid windows, and scores average equally over window centers and RGB channels.
LPIPS~\cite{zhang2018lpips} uses AlexNet~\cite{krizhevsky2012imagenet} and version 0.1, with RGB mapped linearly to $[-1,1]$.
Each metric is evaluated per view, averaged within each scene, and then averaged equally across scenes.
Higher PSNR and SSIM and lower LPIPS indicate closer agreement with the scene-consistent targets.

Temporal residual measures how target-relative color error changes between corresponding scene points in adjacent views.
For each view, the error field is the output's CIELAB value minus its target's CIELAB value at each pixel.
We sort reconstruction views by image name and use the first 32 views, giving 62 directed adjacent-view pairs.
Expected depth and alpha from the original asset establish the correspondences.
For each direction, we reproject the destination-view error field into the source view and average the Euclidean norm of its difference from the source-view error field.
Valid correspondences lie inside the destination image, have positive source depth and projected depth $z$, have source and sampled destination alpha at least $0.5$, and satisfy $|z-d_{v'}|/\max(z,10^{-6})<0.02$, where $d_{v'}$ is sampled destination depth.
Sampling is bilinear at pixel centers with aligned grid corners.
Directions with at least 200 valid pixels contribute equally to the scene score, and scene scores contribute equally within each reported scene group.
Lower temporal residual indicates more consistent target-relative color errors across views.

\noindent\textbf{Statistical Analysis.}

We report population standard deviations across scenes.
Paired operator comparisons use 10,000 percentile-bootstrap replicates over scenes with seed zero.
Each bootstrap replicate contains 37 scenes and preserves the scene-source proportions of \bench.

\begin{table*}[!t]
\centering
\footnotesize
\setlength{\tabcolsep}{3pt}
\begin{tabular}{l|rrrr|rrrr}
\toprule
& \multicolumn{4}{c|}{DL3DV scenes} & \multicolumn{4}{c}{Mip-NeRF 360 scenes} \\
\cmidrule(lr){2-5}\cmidrule(lr){6-9}
Method & \dE$\downarrow$ & PSNR$\uparrow$ & SSIM$\uparrow$ & LPIPS$\downarrow$ & \dE$\downarrow$ & PSNR$\uparrow$ & SSIM$\uparrow$ & LPIPS$\downarrow$ \\
\midrule
\multicolumn{9}{l}{\cellcolor[gray]{0.90}\textit{Aesthetic Color Enhancement}} \\
Original asset & $9.66\pm2.78$ & 22.98 & 0.8958 & 0.0926 & $13.38\pm2.04$ & 19.85 & 0.8412 & 0.1190 \\
Instant Colorization & $22.16\pm5.64$ & 14.04 & 0.6120 & 0.4694 & $27.68\pm4.80$ & 13.03 & 0.4741 & 0.5589 \\
BilaRF & $5.65\pm2.27$ & 27.42 & \textcolor{retouchdeep}{\textbf{0.9402}} & 0.0627 & $8.30\pm4.81$ & 25.59 & \textcolor{retouchdeep}{\textbf{0.9189}} & 0.0843 \\
SAAD & $5.76\pm1.90$ & 27.52 & 0.9131 & 0.0818 & $7.48\pm1.48$ & 25.59 & 0.8918 & 0.0936 \\
ReparamGS & $10.88\pm4.20$ & 19.85 & 0.7234 & 0.2551 & $14.03\pm4.17$ & 17.01 & 0.5996 & 0.3167 \\
FPGS & $21.17\pm5.17$ & 14.76 & 0.6485 & 0.3498 & $22.18\pm3.19$ & 14.66 & 0.5560 & 0.3488 \\
VIRGi & $11.28\pm4.55$ & 20.66 & 0.7758 & 0.2085 & $13.75\pm2.35$ & 18.40 & 0.6527 & 0.2499 \\
ViP3DE & $21.22\pm5.28$ & 14.43 & 0.5656 & 0.5744 & $25.32\pm4.97$ & 13.95 & 0.4243 & 0.5758 \\
\rowcolor{retouchpale}
\method & \textcolor{retouchdeep}{$\mathbf{5.01\pm2.03}$} & \textcolor{retouchdeep}{\textbf{28.64}} & 0.9246 & \textcolor{retouchdeep}{\textbf{0.0537}} & \textcolor{retouchdeep}{$\mathbf{7.04\pm1.78}$} & \textcolor{retouchdeep}{\textbf{26.35}} & 0.9110 & \textcolor{retouchdeep}{\textbf{0.0633}} \\
\midrule
\multicolumn{9}{l}{\cellcolor[gray]{0.90}\textit{Reference-Guided Color Transfer}} \\
Original asset & $9.66\pm2.78$ & 22.98 & 0.8958 & 0.0926 & $13.38\pm2.04$ & 19.85 & 0.8412 & 0.1190 \\
Instant Colorization & $22.06\pm5.64$ & 14.02 & 0.6126 & 0.4666 & $27.37\pm4.80$ & 13.06 & 0.4756 & 0.5561 \\
BilaRF & $5.39\pm2.42$ & 27.99 & \textcolor{retouchdeep}{\textbf{0.9391}} & 0.0632 & $6.78\pm4.71$ & 27.50 & \textcolor{retouchdeep}{\textbf{0.9264}} & 0.0728 \\
SAAD & $14.44\pm4.85$ & 20.46 & 0.8417 & 0.1890 & $18.64\pm5.27$ & 18.96 & 0.7726 & 0.2431 \\
ReparamGS & $10.69\pm4.59$ & 20.12 & 0.7370 & 0.2478 & $13.77\pm4.13$ & 17.17 & 0.5945 & 0.3217 \\
FPGS & $21.28\pm5.34$ & 14.73 & 0.6483 & 0.3507 & $22.70\pm3.13$ & 14.55 & 0.5520 & 0.3543 \\
VIRGi & $11.10\pm4.50$ & 20.80 & 0.7765 & 0.2069 & $13.37\pm2.37$ & 18.59 & 0.6540 & 0.2492 \\
ViP3DE & $21.19\pm5.21$ & 14.38 & 0.5645 & 0.5773 & $26.12\pm5.70$ & 13.77 & 0.4158 & 0.5900 \\
\rowcolor{retouchpale}
\method & \textcolor{retouchdeep}{$\mathbf{4.05\pm1.57}$} & \textcolor{retouchdeep}{\textbf{29.56}} & 0.9229 & \textcolor{retouchdeep}{\textbf{0.0541}} & \textcolor{retouchdeep}{$\mathbf{3.99\pm0.93}$} & \textcolor{retouchdeep}{\textbf{29.26}} & 0.9193 & \textcolor{retouchdeep}{\textbf{0.0491}} \\
\bottomrule
\end{tabular}
\caption{Color and image fidelity by scene source within \bench. Color error reports mean and population standard deviation. Other metrics report scene means. Bold marks the best result within each task.}
\label{tab:extended_nr}
\label{tab:extended_ref1}
\end{table*}

\begin{table*}[!t]
    \centering
    \small
    \setlength{\tabcolsep}{6pt}
    \begin{tabular}{l|rr|rr|rr|rr}
        \toprule
        & \multicolumn{4}{c|}{DL3DV scenes} & \multicolumn{4}{c}{Mip-NeRF 360 scenes} \\
        \cmidrule(lr){2-5}\cmidrule(lr){6-9}
        & \multicolumn{2}{c|}{Enhancement} & \multicolumn{2}{c|}{Transfer} & \multicolumn{2}{c|}{Enhancement} & \multicolumn{2}{c}{Transfer} \\
        \cmidrule(lr){2-3}\cmidrule(lr){4-5}\cmidrule(lr){6-7}\cmidrule(lr){8-9}
        Method & Gap & Temporal$\downarrow$ & Gap & Temporal$\downarrow$ & Gap & Temporal$\downarrow$ & Gap & Temporal$\downarrow$ \\
        \midrule
        BilaRF & $+2.18$ & \textcolor{retouchdeep}{\textbf{0.71}} & $+2.55$ & \textcolor{retouchdeep}{\textbf{0.72}} & $+3.44$ & \textcolor{retouchdeep}{\textbf{2.20}} & $+3.72$ & \textcolor{retouchdeep}{\textbf{1.95}} \\
        SAAD & $+1.57$ & 0.90 & $+11.62$ & 1.24 & $+1.42$ & 3.34 & $+14.85$ & 4.33 \\
        ReparamGS & $+3.16$ & 2.20 & $+3.30$ & 2.13 & $+3.74$ & 7.05 & $+4.16$ & 7.15 \\
        VIRGi & $+4.46$ & 1.59 & $+4.62$ & 1.59 & $+4.95$ & 5.39 & $+5.62$ & 5.37 \\
        ViP3DE & $+9.83$ & 2.97 & $+10.75$ & 3.09 & $+11.41$ & 9.96 & $+11.32$ & 9.58 \\
        \midrule
        \rowcolor{retouchpale}
        \method & $+0.10$ & 0.73 & $+0.18$ & 0.75 & $-0.80$ & 2.36 & $+0.35$ & 2.20 \\
        \bottomrule
    \end{tabular}
    \caption{View-generalization gap and temporal residual by scene source within \bench. Gap is mean off-trajectory \dE minus reference-view \dE, with positive values indicating increased error at novel views. Temporal residual follows the correspondence evaluation in \cref{app:aux_metrics}. Bold marks the lowest temporal residual in each setting.}
    \label{tab:extended_view_gap}
\end{table*}

\section{Extended Experimental Results}
\label{app:extended_results}

\subsection{Color and Image Fidelity}
\label{app:extended_metrics}

\begin{figure*}[p]
\centering
\includegraphics[width=\textwidth]{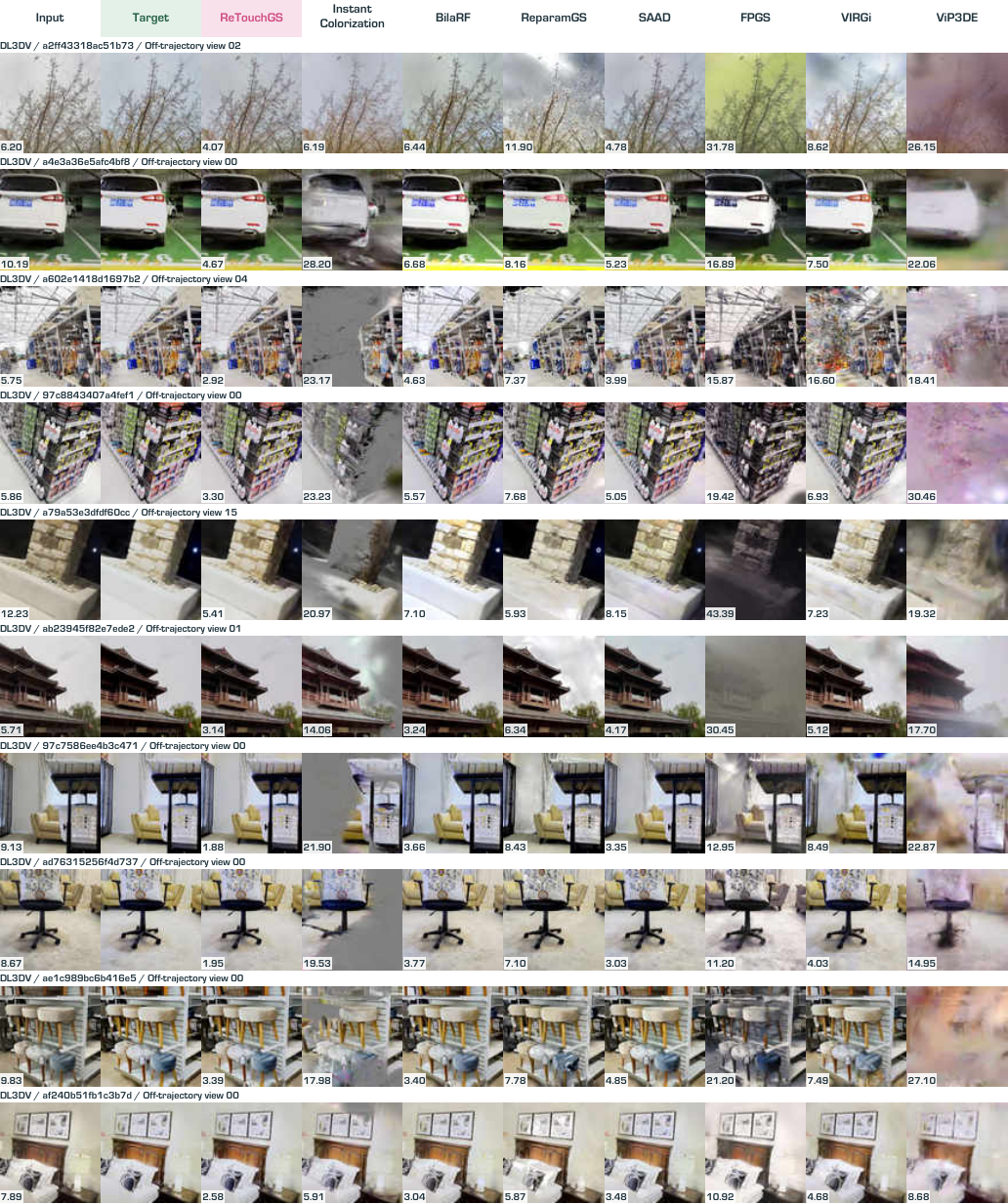}
\caption{Additional aesthetic color enhancement on off-trajectory views. Each row compares the same view and image region. Numbers report mean pixelwise \dE$\downarrow$ over the displayed region.}
\label{fig:app_enhancement_scenes}
\end{figure*}

\begin{figure*}[p]
\centering
\includegraphics[width=\textwidth]{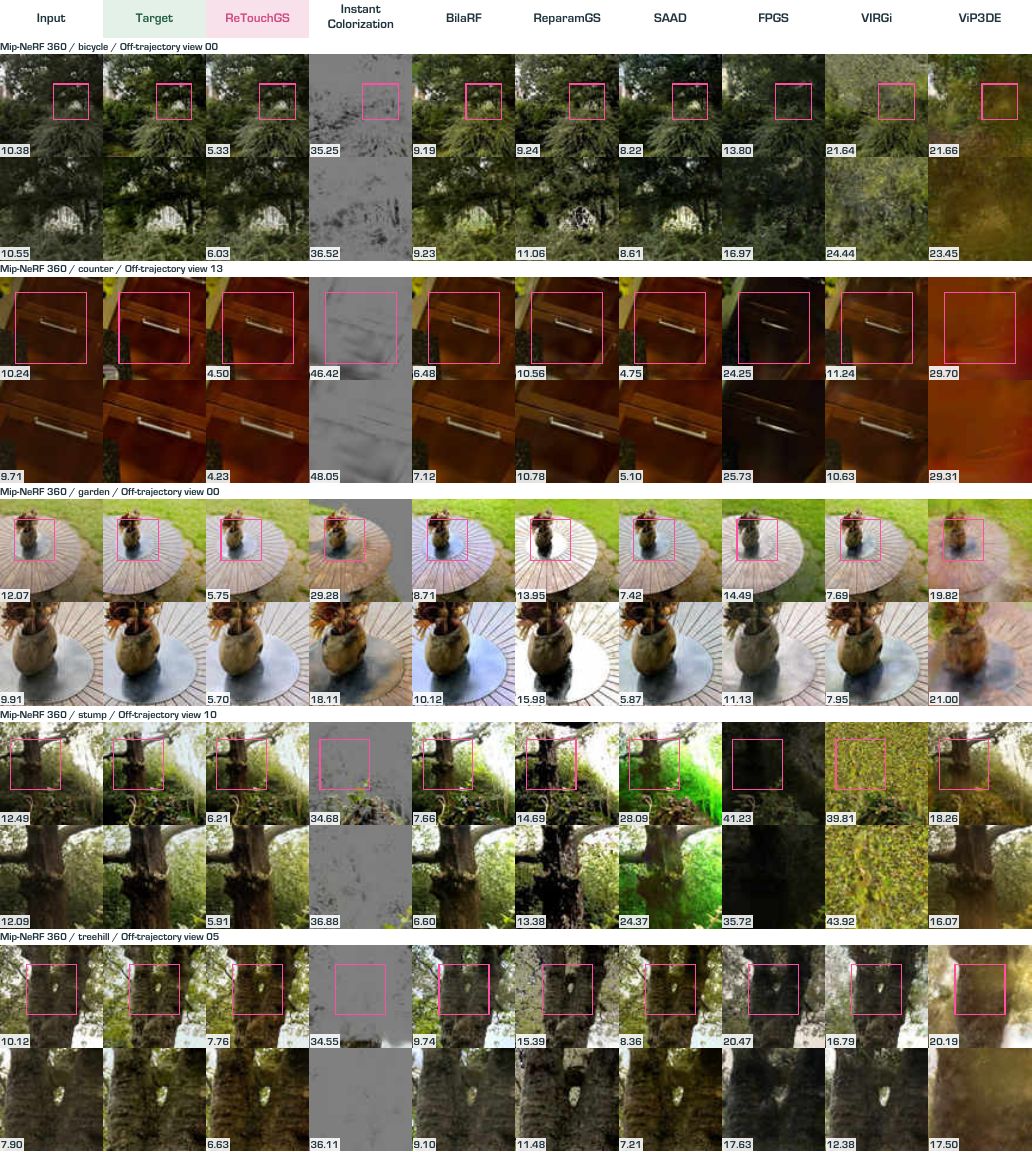}
\caption{Aesthetic color enhancement on off-trajectory views with matched detail crops. Each overview is followed by the same enlarged region across methods. Numbers report mean pixelwise \dE$\downarrow$ over the displayed region.}
\label{fig:app_enhancement_details}
\end{figure*}

\Cref{tab:extended_nr} breaks down the main asset comparisons by scene source within \bench.
\method achieves the lowest color error and LPIPS and the highest PSNR in both tasks on both source subsets of \bench.
These results support accurate color reproduction across the diverse assets in \bench.
Alternative LUT configurations are analyzed in \cref{app:operator_transfer}.
\Cref{fig:app_enhancement_scenes,fig:app_enhancement_details} extend the aesthetic enhancement comparisons to additional scenes and matched detail crops.
The examples show how \method adjusts colors and tones across vegetation, reflective surfaces, and furnished interiors while preserving scene structure.
Numbers in the additional qualitative figures report mean pixelwise \dE over the displayed eight-bit image regions.
All output columns within a comparison row share the same viewpoint and image region.

\begin{figure*}[p]
\centering
\includegraphics[width=\textwidth]{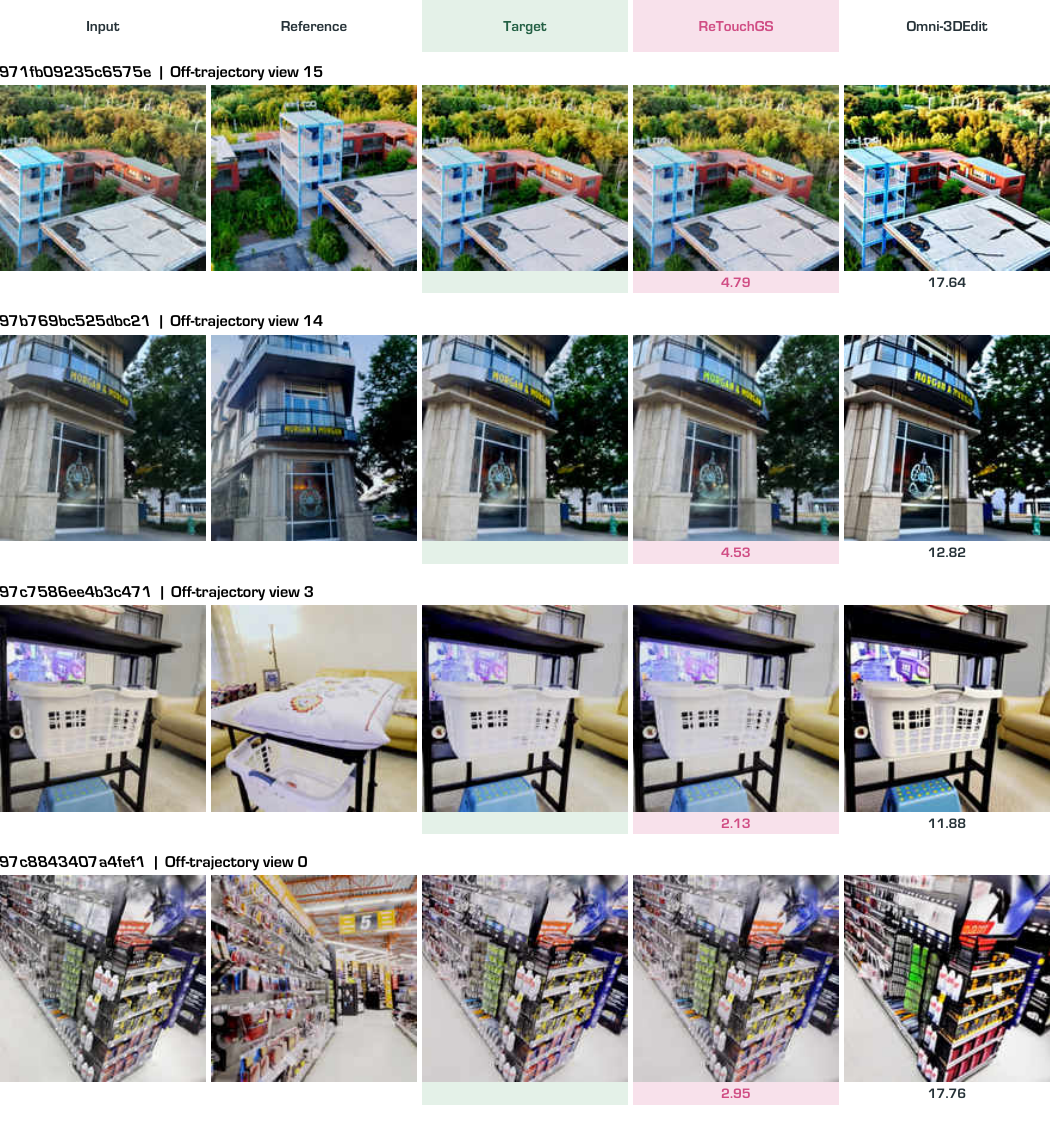}
\caption{Reference-guided color transfer compared with Omni-3DEdit. Numbers report mean pixelwise \dE$\downarrow$.}
\label{fig:app_transfer_image_comparison}
\end{figure*}

\begin{figure*}[p]
\centering
\includegraphics[width=\textwidth]{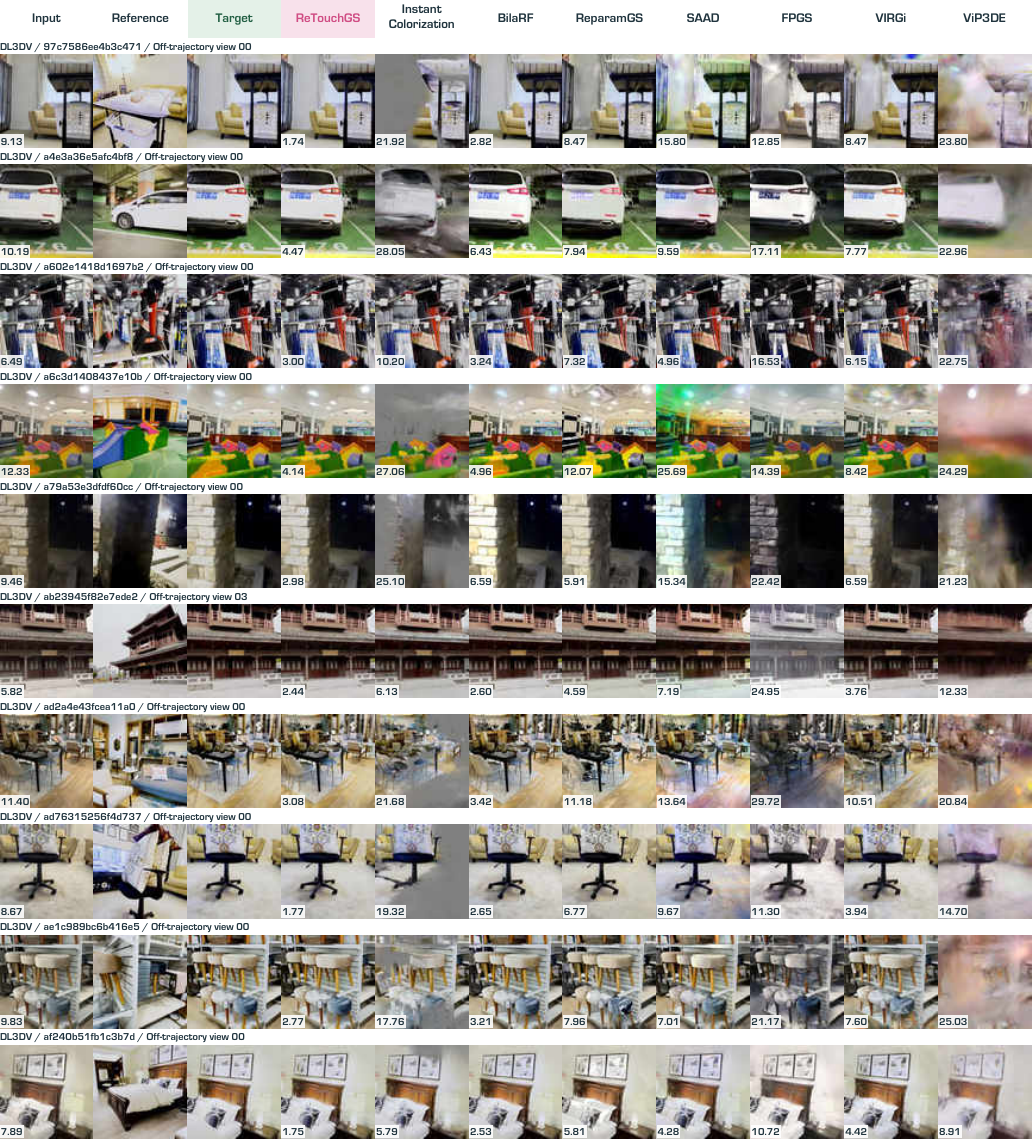}
\caption{Additional reference-guided color transfer with GLUT on off-trajectory views. Reference is the edited input view. All output columns share the evaluation view and region. Numbers report mean pixelwise \dE$\downarrow$ over the displayed region.}
\label{fig:app_transfer_glut_scenes}
\end{figure*}

\begin{figure*}[p]
\centering
\includegraphics[width=\textwidth]{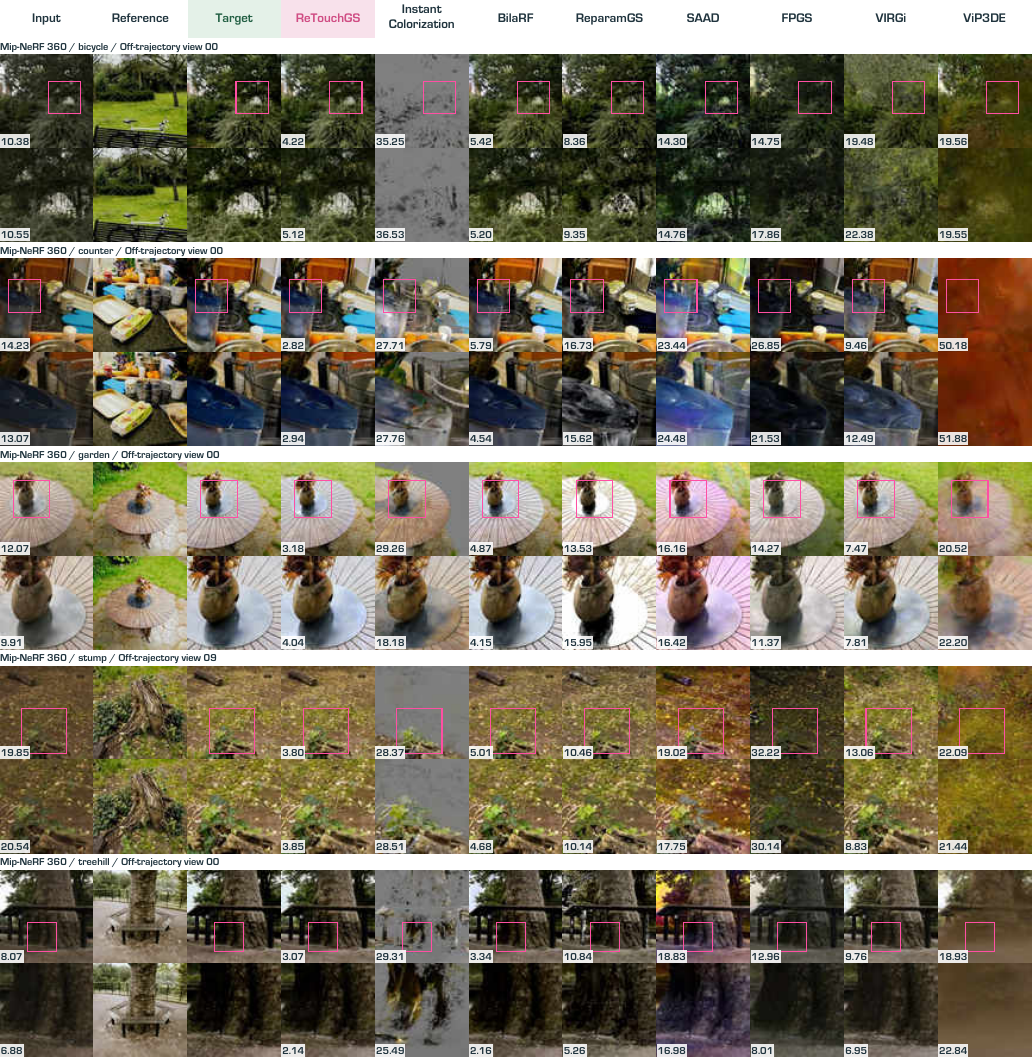}
\caption{Reference-guided color transfer with GLUT on off-trajectory views and matched detail crops. Each overview is followed by the same enlarged region across methods. Numbers report mean pixelwise \dE$\downarrow$ over the displayed region.}
\label{fig:app_transfer_glut_details}
\end{figure*}

\Cref{fig:app_transfer_image_comparison} compares images edited by Omni-3DEdit~\cite{chen2026omni3dedit} with renders from \method.
\method produces coherent color changes across buildings, indoor surfaces, and small objects in this comparison.
\Cref{fig:app_transfer_glut_scenes,fig:app_transfer_glut_details} show reference colors reproduced across broad surfaces and fine structures in additional scenes.



\begin{table*}[!t]
    \centering
    \footnotesize
    \setlength{\tabcolsep}{3pt}
    \begin{tabular}{ll|r|rrr|rrr}
        \toprule
        & & Operator & \multicolumn{3}{c|}{Asset error $\downarrow$} & \multicolumn{3}{c}{Operator fidelity error $\downarrow$} \\
        \cmidrule(lr){4-6}\cmidrule(lr){7-9}
        Scene source & Operator source & error $\downarrow$ & Direct & Analytic & Refined & Direct & Analytic & Refined \\
        \midrule
        DL3DV & Training mean & 4.89 & 7.01 & 6.52 & 6.23 & 4.51 & 3.82 & 3.30 \\
        \rowcolor{retouchpale}
        & Aesthetic enhancement network & 2.99 & 6.04 & 5.48 & 5.01 & 4.94 & 4.24 & 3.66 \\
        & Image predictor & 2.10 & 5.51 & 4.94 & 4.41 & 5.01 & 4.29 & 3.68 \\
        & Enhancer $\times1$ & 2.46 & 5.69 & 5.39 & 4.84 & 5.12 & 4.70 & 4.05 \\
        & Enhancer $\times4$ & 1.49 & 5.35 & 5.00 & 4.36 & 5.27 & 4.91 & 4.21 \\
        & Enhancer $\times20$ & 0.52 & 5.18 & 4.73 & 4.06 & 5.18 & 4.74 & 4.06 \\
        & Reference-fitted LUT & 0.33 & 5.26 & 4.68 & 4.05 & 5.06 & 4.52 & 3.90 \\
        \midrule
        Mip-NeRF 360 & Training mean & 10.34 & 10.72 & 10.42 & 10.52 & 4.34 & 3.36 & 2.92 \\
        \rowcolor{retouchpale}
        & Aesthetic enhancement network & 6.07 & 7.58 & 7.26 & 7.04 & 4.84 & 3.97 & 3.43 \\
        & Image predictor & 4.96 & 6.93 & 6.68 & 6.18 & 5.11 & 4.34 & 3.72 \\
        & Enhancer $\times1$ & 4.70 & 6.95 & 6.51 & 6.28 & 5.39 & 4.37 & 3.83 \\
        & Enhancer $\times4$ & 3.29 & 5.98 & 5.73 & 5.30 & 5.26 & 4.63 & 3.94 \\
        & Enhancer $\times20$ & 0.97 & 5.16 & 4.58 & 3.94 & 5.10 & 4.46 & 3.81 \\
        & Reference-fitted LUT & 0.53 & 5.41 & 4.56 & 3.99 & 5.04 & 4.26 & 3.68 \\
        \bottomrule
    \end{tabular}
    \caption{Color-operator accuracy and color-update fidelity by scene source within \bench. All errors are \dE. Direct, Analytic, and Refined denote the three configurations defined in the text.}
    \label{tab:app_operator_decomposition}
\end{table*}

\subsection{Accuracy Across Viewpoint Groups}
\label{app:extended_views}
\label{app:view_results}

The small view-generalization gaps of \method are accompanied by low temporal residuals in both tasks (\cref{tab:extended_view_gap}).
The latter evaluates target-relative errors at corresponding scene points, complementing the viewpoint-group comparison.
These measurements support a consistent realization of the desired color adjustment across viewpoints.
\Cref{fig:app_enhancement_multiview} shows held-out views rendered from one enhanced asset per scene.
\Cref{fig:app_aligned_multiview} extends this comparison to both tasks with five aligned held-out viewpoints per scene.
The shared color update preserves the appearance of outdoor structures and indoor surfaces across viewpoints, while the matched crops relate the resulting colors directly to the target.


\begin{table*}[!t]
    \centering
    \small
    \setlength{\tabcolsep}{5pt}
    \begin{tabular}{lll|rrrr|r}
        \toprule
        Scene source & Operator & Color update & \dE$\downarrow$ & PSNR$\uparrow$ & SSIM$\uparrow$ & LPIPS$\downarrow$ & Fidelity error$\downarrow$ \\
        \midrule
        DL3DV & Grid LUT & Analytic & 4.679 & 28.41 & 0.9145 & 0.0613 & 4.518 \\
        \rowcolor{retouchpale}
        & Grid LUT & Refined & 4.049 & 29.56 & 0.9229 & 0.0541 & 3.901 \\
        & GLUT & Analytic & 4.406 & 28.90 & 0.9203 & 0.0542 & 4.230 \\
        \rowcolor{retouchpale}
        & GLUT & Refined & \textcolor{retouchdeep}{\textbf{3.851}} & \textcolor{retouchdeep}{\textbf{29.93}} & \textcolor{retouchdeep}{\textbf{0.9264}} & \textcolor{retouchdeep}{\textbf{0.0493}} & \textcolor{retouchdeep}{\textbf{3.669}} \\
        \midrule
        Mip-NeRF 360 & Grid LUT & Analytic & 4.556 & 28.37 & 0.9071 & 0.0576 & 4.262 \\
        \rowcolor{retouchpale}
        & Grid LUT & Refined & 3.990 & 29.26 & 0.9193 & 0.0491 & 3.684 \\
        & GLUT & Analytic & 4.214 & 29.67 & 0.9173 & 0.0485 & 3.859 \\
        \rowcolor{retouchpale}
        & GLUT & Refined & \textcolor{retouchdeep}{\textbf{3.761}} & \textcolor{retouchdeep}{\textbf{30.37}} & \textcolor{retouchdeep}{\textbf{0.9252}} & \textcolor{retouchdeep}{\textbf{0.0440}} & \textcolor{retouchdeep}{\textbf{3.415}} \\
        \bottomrule
    \end{tabular}
    \caption{Color-operator and refinement combinations for reference-guided color transfer. Each row uses all scenes from its source group and the same sixteen off-trajectory viewpoints. Fidelity error denotes operator fidelity error, measured by \dE relative to the input operator applied to the original render.}
    \label{tab:app_glut_cross}
\end{table*}

\section{Color Operator and Refinement Analysis}
\label{app:refinement_analysis}

\subsection{Color Operators and Refinement}
\label{app:exp_operators}
\label{app:operator_analysis}

We separate color-operator accuracy from its realization in the asset.
Operator error measures how closely the input color operator's output $T(\Pi(R(\mathcal G,v)))$ matches the benchmark target $I_s^\star(v)$.
Asset error compares the edited render $R(\mathcal G',v)$ with the same target.
Operator fidelity error compares the edited render with $T(\Pi(R(\mathcal G,v)))$, measuring how accurately the color update realizes the input color operator.
All use \dE on the sixteen off-trajectory views, averaged first within each scene and then across scenes.
\Cref{tab:app_operator_decomposition} compares seven grid-LUT sources with direct per-Gaussian transformation, the analytic color update, and learned refinement.
For each operator source, the three color-update configurations share the operator, original asset, and viewpoints.
Direct corresponds to removing blended color context and refinement in \cref{tab:exp_ablation}. Analytic retains the analytic color update and removes the blend-aware color refinement network. Refined adds the refinement network to the analytic color update, matching the color-update configuration of Full model in that table.
The training-mean LUT is fixed across test scenes, while the aesthetic enhancement network predicts a LUT from each asset and its associated camera parameters.
The image predictor reads twenty original renders, and the enhancer-based controls fit their LUTs to one, four, or twenty enhanced input views.
The reference-fitted LUT uses the reference image pair.

The blend-aware color refinement network reduces operator fidelity error for all 259 scene--operator pairs across the seven operator sources in \bench.
This demonstrates that the blend-aware color refinement network accurately realizes color mappings obtained through different LUT prediction and fitting procedures.
Relative to the analytic color update, it also improves PSNR in 240 pairs, SSIM in 252, and LPIPS in 251.
For the training-mean LUT on Mip-NeRF 360, the refined asset follows the supplied operator more closely, with operator fidelity error decreasing from 3.36 to 2.92, while its benchmark-target error changes from 10.42 to 10.52.
These measurements distinguish selecting a mapping that matches the target from accurately realizing that mapping in 3D.

\subsection{Alternative Color Operators}
\label{app:exp_transfer}
\label{app:operator_transfer}

\noindent{\textbf{Alternative LUT Representation: GLUT.}}
We use grid LUTs in the main experiments to maintain a consistent color-operator representation across both tasks and the color-update ablations.
As a supplementary evaluation, we replace the grid LUT with GLUT~\cite{xue2026glut} for reference-guided color transfer, using the same reference pair and the fitting procedures in \cref{app:alternative_operators}.

\Cref{tab:app_glut_cross} shows that blend-aware color refinement improves all four asset metrics for both operators in both source subsets of \bench.
The grid LUT has lower operator error on the DL3DV and Mip-NeRF 360 scenes, at 0.333 and 0.529~\dE, respectively, compared with 0.781 and 1.088 for GLUT.
GLUT gives lower asset error with both color-update procedures, illustrating the joint role of the color mapping and Gaussian blending.
Adding blend-aware color refinement to the analytic color update reduces asset error in all 37 scenes for each operator.
The mean paired reduction is 0.614~\dE for the grid LUT, with a 95\% confidence interval of $[0.518,0.715]$, and 0.530 for GLUT, with $[0.433,0.644]$.
The bootstrap procedure is specified in \cref{app:aux_metrics}.
With blend-aware color refinement, GLUT improves asset error in 36 of 37 scenes relative to the grid LUT.
Averaged equally over all 37 scenes, replacing the grid LUT with GLUT further reduces \dE from 4.035 to 3.829 and LPIPS from 0.0529 to 0.0480, while increasing PSNR from 29.49 to 30.04 and SSIM from 0.9221 to 0.9261.
These results show that the same blend-aware color refinement network can benefit from different LUT representations.

\noindent{\textbf{Alternative LUT Source: VCG.}}
\Cref{tab:operator_replacement} compares Video Color Grading (VCG)~\cite{video_color_grading} with our default grid LUT configurations.
With the analytic color update fixed, the default LUTs achieve lower color error than VCG-generated LUTs in both tasks.
Adding blend-aware color refinement further improves the grid-LUT results.
The LUT prediction and fitting procedures are specified in \cref{app:alternative_operators,app:operator_fitting}.

\begin{table}[!htb]
\centering
\footnotesize
\setlength{\tabcolsep}{5pt}
\resizebox{\columnwidth}{!}{%
\begin{tabular}{l|rr|rr}
\toprule
& \multicolumn{2}{c|}{Enhancement} & \multicolumn{2}{c}{Transfer} \\
\cmidrule(lr){2-3}\cmidrule(lr){4-5}
Operator + update & \dE$\downarrow$ & PSNR$\uparrow$ & \dE$\downarrow$ & PSNR$\uparrow$ \\
\midrule
VCG + analytic & 8.388 & 24.95 & 8.322 & 25.03 \\
Grid LUT + analytic & 5.911 & 27.37 & 4.649 & 28.40 \\
\midrule
\rowcolor{retouchpale}
Grid LUT + refined & \textcolor{retouchdeep}{\textbf{5.502}} & \textcolor{retouchdeep}{\textbf{28.08}} & \textcolor{retouchdeep}{\textbf{4.035}} & \textcolor{retouchdeep}{\textbf{29.49}} \\
\bottomrule
\end{tabular}%
}
\caption{LUT configuration comparisons averaged equally over 37 scenes.}
\label{tab:operator_replacement}
\end{table}

\begin{table*}[!t]
    \centering
    \small
    \setlength{\tabcolsep}{6pt}
    \begin{tabular}{lr|rr|rr}
        \toprule
        & & \multicolumn{2}{c|}{NamedCurves MIT5K-DPE} & \multicolumn{2}{c}{Image-adaptive 3D LUT} \\
        \cmidrule(lr){3-4}\cmidrule(lr){5-6}
        Predictor input & Training scenes & Validation & Held-out views & Validation & Held-out views \\
        \midrule
        Original renders & 36 & 3.525 & 3.439 & 2.201 & 2.573 \\
        Original renders & 72 & 2.858 & 2.661 & 1.665 & 1.890 \\
        Original renders & 145 & 2.464 & 2.327 & 1.298 & 1.656 \\
        Original renders & 189 & 2.324 & 2.124 & 1.292 & 1.561 \\
        \midrule
        Projected CNN & 145 & 3.522 & 3.203 & 1.830 & 1.881 \\
        Color quantiles & 145 & 3.604 & 3.620 & 1.944 & 2.198 \\
        Projected CNN + quantiles & 145 & 3.086 & 2.952 & 1.816 & 1.957 \\
        \rowcolor{retouchpale}
        Projected CNN + quantiles & 189 & 3.062 & 3.008 & 1.662 & 1.871 \\
        \bottomrule
    \end{tabular}
    \caption{Operator prediction before the color update, measured by \dE$\downarrow$ over all 23 validation scenes and all 28 DL3DV test scenes. All predictors use twenty input viewpoints. Errors are measured on input views for validation and sixteen held-out views for testing.}
    \label{tab:app_predictor}
\end{table*}

\subsection{Aesthetic Operator Prediction}
\label{app:exp_prediction}

To test whether spatial appearance and the overall color distribution provide complementary cues for render-free LUT prediction, we ablate the projected CNN features and color quantiles separately.
We also compare with predictors driven by original renders to examine prediction accuracy across input representations and training-set sizes.

Every predictor decodes a $17^3$ LUT from 32 PCA coefficients using a basis fitted to its own training targets.
Image predictors read twenty original renders at width 240.
Variants of the aesthetic enhancement network construct width-80 projection grids directly from Gaussian attributes and associated camera parameters.
The projected-CNN variant omits the color quantiles, and the quantile-only variant zeros the pooled CNN features.
The combined network concatenates both inputs.

All variants follow the aesthetic enhancement network's optimization and model-selection settings in \cref{app:training}.
Image predictors sample four to twenty views per training scene and supervise image pixels.
Aesthetic enhancement network variants sample four to eight viewpoints and use the color bank $\mathcal X_s$ for supervision.
The 36- and 72-scene training sets are nested subsets of the 145-scene training set, sampled with seed zero.
The 189-scene training set contains these 145 scenes and 44 further scenes.

\Cref{tab:app_predictor} measures operator error by comparing predicted and target LUT outputs on the same original renders.
Validation uses twenty input views at width 240 per scene. Testing samples 30,000 pixels from each of sixteen native-resolution held-out views per scene.
For NamedCurves~\cite{serrano2024namedcurves}, combining projected appearance and color quantiles improves both validation and held-out-view prediction over either input alone.
Increasing the training set to 189 scenes improves validation error for both retouching priors.

At matched training-set sizes, image predictors achieve lower operator error (\cref{tab:app_predictor}).
This advantage is consistent with their access to rendered appearance, which explicitly captures visibility and blended pixel colors.
Our aesthetic enhancement network summarizes scene appearance through coarse Gaussian projections and color quantiles, enabling direct prediction from the asset without rendering input images.
Both predictor types learn the retouching prior during training and require no test-time image-enhancer calls.
The comparison reflects their respective input representations, resolutions, and supervision schemes.

\subsection{Refinement Mechanisms and Training}
\label{app:exp_refinement}
\label{app:training_analysis}

\begin{figure*}[p]
\centering
\includegraphics[width=\textwidth]{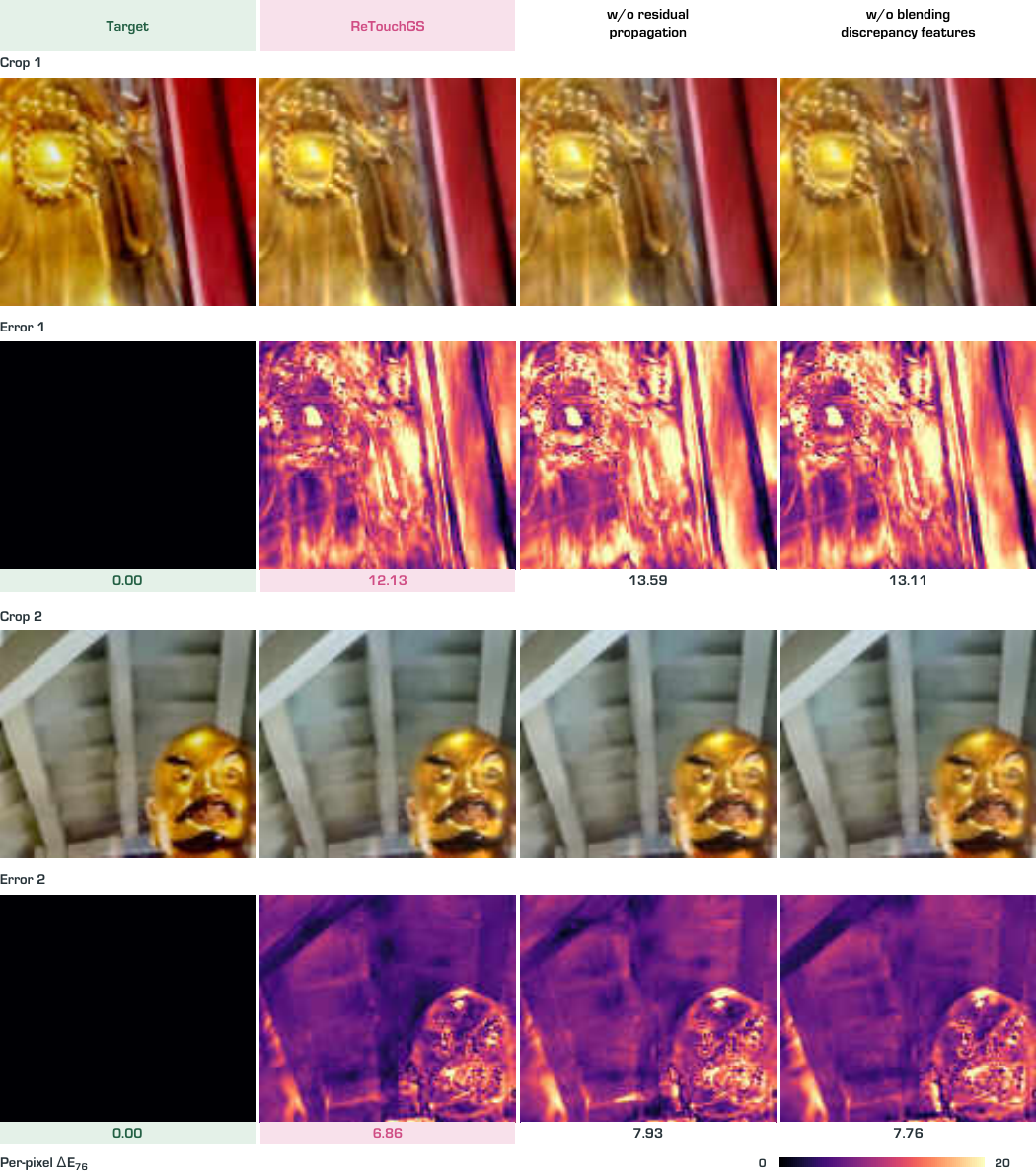}
\caption{Local effects of refinement mechanisms for aesthetic color enhancement. Matched crops and pixelwise error maps compare the full model with two component ablations. All variants share the grid LUT. Error maps use a common \dE scale from 0 to 20.}
\label{fig:app_enhancement_refinement}
\end{figure*}

\begin{figure*}[p]
\centering
\includegraphics[width=\textwidth]{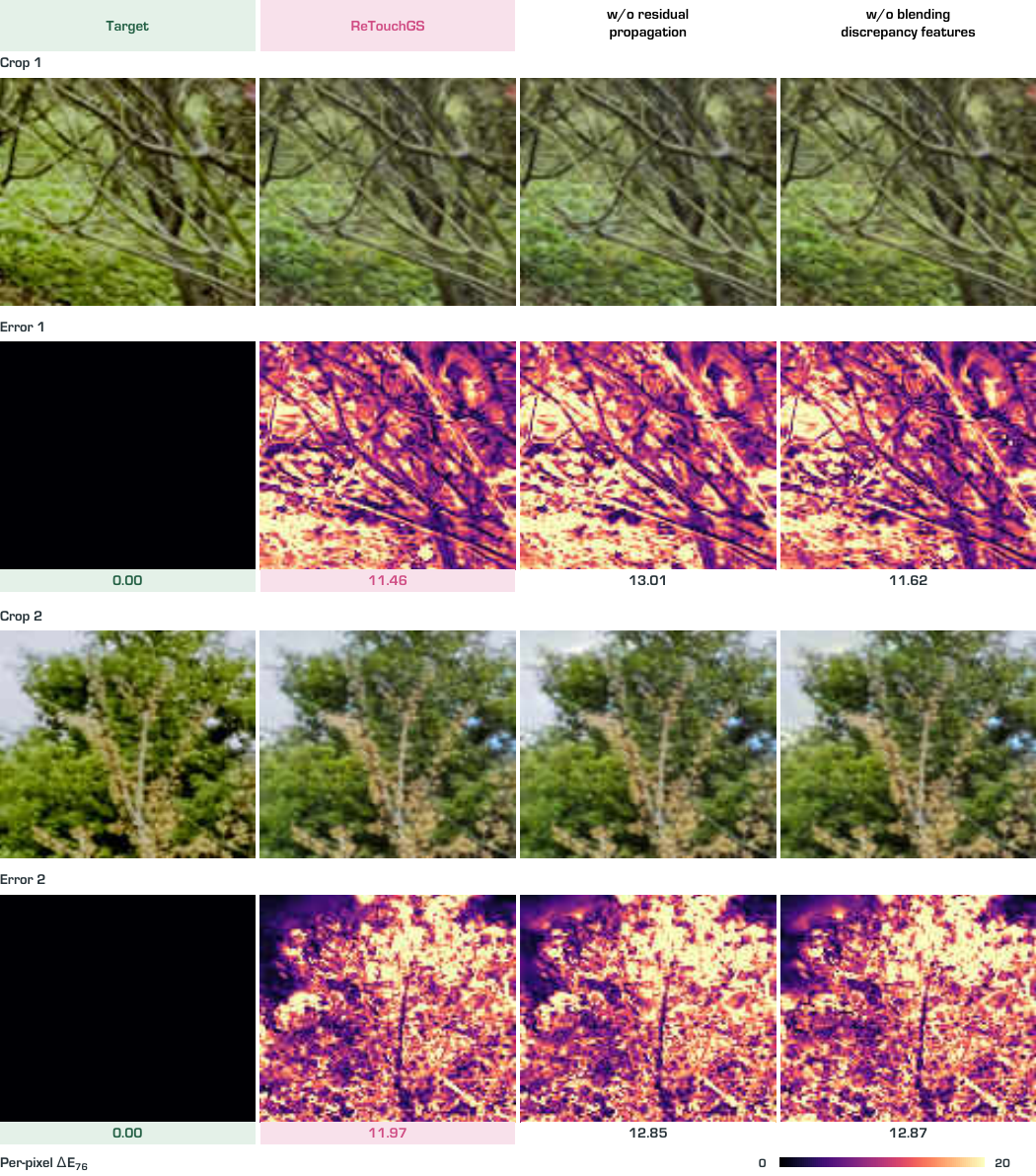}
\caption{Local effects of refinement mechanisms for reference-guided color transfer. Matched crops and pixelwise error maps compare the full model with two component ablations. All variants share the grid LUT. Error maps use a common \dE scale from 0 to 20.}
\label{fig:app_transfer_refinement}
\end{figure*}

The refinement mechanism ablations on all 37 scenes share training scenes, solver demonstrations, and rendering fine-tuning steps.
The residual-propagation ablation removes cross-Gaussian residual contributions and edge blending discrepancy contributions while retaining self residual updates and neighbor feature aggregation.
\Cref{tab:app_ablation_datasets} breaks down the scene-weighted means in \cref{tab:exp_ablation} by scene source.
\Cref{fig:app_enhancement_refinement,fig:app_transfer_refinement} isolate the local effects of residual propagation and blending discrepancy features in the two tasks.
All variants use the same grid LUT and viewpoint.
The matched crops and error maps show that both mechanisms reduce local color deviations around reflective details and thin foliage, connecting the aggregate gains to spatially localized corrections.

\begin{table}[!htb]
\centering
\footnotesize
\setlength{\tabcolsep}{1.5pt}
\resizebox{\columnwidth}{!}{%
\begin{tabular}{l|rr|rr}
\toprule
& \multicolumn{2}{c|}{DL3DV} & \multicolumn{2}{c}{Mip-NeRF 360} \\
\cmidrule(lr){2-3}\cmidrule(lr){4-5}
Configuration & Enhancement & Transfer & Enhancement & Transfer \\
\midrule
\multicolumn{5}{l}{\cellcolor[gray]{0.90}\textit{Color update components}} \\
w/o blended color context \& refinement & 6.043 & 5.264 & 7.580 & 5.406 \\
w/o blend-aware color refinement & 5.477 & 4.679 & 7.261 & 4.556 \\
\multicolumn{5}{l}{\cellcolor[gray]{0.90}\textit{Refinement mechanisms}} \\
w/o residual propagation & 5.129 & 4.179 & \textcolor{retouchdeep}{\textbf{7.038}} & 4.220 \\
w/o blending discrepancy features & 5.087 & 4.151 & 7.044 & 4.134 \\
\midrule
\rowcolor{retouchpale}
Full model & \textcolor{retouchdeep}{\textbf{5.007}} & \textcolor{retouchdeep}{\textbf{4.049}} & 7.042 & \textcolor{retouchdeep}{\textbf{3.990}} \\
\bottomrule
\end{tabular}%
}
\caption{Color update ablations by scene source (\dE$\downarrow$). Each configuration uses all 37 scenes in \bench with fixed grid LUTs.}
\label{tab:app_ablation_datasets}
\end{table}

The training-strategy comparison uses six validation scenes and four operators per scene from NamedCurves~\cite{serrano2024namedcurves} MIT5K-DPE, PPR10K-C, MIT5K-UEGAN, and image-adaptive 3D LUT~\cite{zeng2020learning}.
Each scene score averages its four operators before computing the mean and population standard deviation across scenes.
We compare two-stage training with rendering-only training at matched rendering and total update budgets.
Rendering-only training uses either 12,000 updates to match the rendering budget or 28,000 updates to match the total training budget.
Solver distillation and rendering-only training use a peak learning rate of $10^{-3}$.
The rendering fine-tuning stage uses $3\times10^{-4}$.
Two-stage training achieves $2.629\pm0.530$~\dE, compared with $2.666\pm0.542$ for 12,000 rendering-only updates and $2.657\pm0.541$ for 28,000 updates.
It improves over the matched-total-budget alternative in all six scenes, supporting solver distillation as an effective initialization for rendering fine-tuning.

\subsection{Target and Visibility Analysis}
\label{app:exp_target_visibility}

\begin{table*}[!t]
\centering
\footnotesize
\setlength{\tabcolsep}{4pt}
\begin{tabular}{ll|rr|rr|rrr}
\toprule
& & \multicolumn{2}{c|}{Scene-consistent target} & \multicolumn{2}{c|}{Per-view target} & \multicolumn{3}{c}{Unobserved regions} \\
\cmidrule(lr){3-4}\cmidrule(lr){5-6}\cmidrule(lr){7-9}
Scene source & Task & Analytic & Refined & Analytic & Refined & Direct & Analytic & Refined \\
\midrule
DL3DV & Enhancement & 5.48 & 5.01 & 6.34 & 5.90 & 6.662 & 6.564 & \textcolor{retouchdeep}{\textbf{6.301}} \\
& Transfer & 4.68 & 4.05 & 5.76 & 5.23 & 5.992 & 6.173 & \textcolor{retouchdeep}{\textbf{5.530}} \\
\midrule
Mip-NeRF 360 & Enhancement & 7.26 & 7.04 & 9.33 & 9.19 & 7.521 & 7.480 & \textcolor{retouchdeep}{\textbf{7.045}} \\
& Transfer & 4.56 & 3.99 & 7.31 & 6.98 & 5.383 & 5.309 & \textcolor{retouchdeep}{\textbf{4.623}} \\
\bottomrule
\end{tabular}
\caption{Target and visibility analyses (\dE$\downarrow$). Target comparisons average scenes equally. Unobserved-region scores pool selected pixels as defined in the text.}
\label{tab:app_target_sensitivity}
\label{tab:app_unobserved}
\end{table*}

\noindent{\textbf{Scene-Consistent and Per-View Targets.}}
We also evaluate edited assets against NamedCurves~\cite{serrano2024namedcurves} outputs produced independently at each off-trajectory viewpoint.
The enhancer receives the original render quantized to eight bits, and its output is quantized to eight bits for scoring.
\Cref{tab:app_target_sensitivity} uses the grid-LUT configuration for both tasks, holding each input operator fixed across color-update procedures.
Blend-aware color refinement improves mean accuracy under both target definitions for both tasks in both source subsets of \bench.
The per-view targets include the image enhancer's view-dependent adjustments, while the scene-consistent targets evaluate consistent colors and tones throughout the asset.

\noindent{\textbf{Regions Beyond the Input Views.}}
We identify Gaussians whose accumulated rendered compositing weight over the twenty input viewpoints is below $10^{-6}$.
At each off-trajectory viewpoint, we select foreground pixels with accumulated alpha above 0.5 for which these Gaussians contribute more than half of that alpha.
The selected pixels occupy 2.37\% of the pooled foreground on DL3DV and 4.43\% on Mip-NeRF 360.
The masks come from the original asset and remain fixed across methods.
\Cref{tab:app_unobserved} pools color-error sums and selected-pixel counts across views and scenes.
Learned refinement improves over the analytic color update in all four settings, supporting color transfer into regions weakly covered by the input views.

\begin{table*}[!t]
\centering
\footnotesize
\setlength{\tabcolsep}{3pt}
\begin{tabular}{l|rrr|rrr|rl}
\toprule
& \multicolumn{3}{c|}{Aesthetic Color Enhancement} & \multicolumn{3}{c|}{Reference-Guided Color Transfer} & & \\
\cmidrule(lr){2-4}\cmidrule(lr){5-7}
Method & Time (s) & Network (M) & Edit vars. & Time (s) & Network (M) & Edit vars. & Prep. vars. & GPU \\
\midrule
Instant Colorization & 10.95 & 0.9936 & 82.24M & 8.98 & 0 & 82.24M & 0 & RTX 5090 \\
BilaRF & 122.54 & 0.994 & 381 & 121.32 & 0.000381 & 381 & 0 & RTX 5090 \\
SAAD & 3,461.78 & 1.008 & 14.56K & 340.57 & 0.01456 & 14.56K & 0 & RTX 5090 \\
ReparamGS & 136.08 & 0.9936 & 612 & 6.02 & 0 & 612 & 164.5M & RTX 5090 \\
FPGS & 633.82 & 31.3 & 0 & 17.41 & 30.3 & 0 & 119.1M & RTX 5090 \\
VIRGi & 22,993.69 & 13.68 & 6.465K & 76.60 & 12.68 & 6.465K & 12.68M & RTX 5090 \\
ViP3DE & 279.26 & 3040 & 100.9M & 277.93 & 3039 & 100.9M & 0 & RTX 5090 \\
Omni-3DEdit & 173.92 & 4868 & 0 & 153.98 & 4867 & 0 & 0 & RTX 5090 \\
\midrule
\rowcolor{retouchpale}
\method & 22.97 & 0.1802 & 0 & 19.32 & 0.04472 & 14.74K & 0 & RTX 5090 \\
\bottomrule
\end{tabular}
\caption{Computational cost over 37 scenes. Times and scene-dependent variable counts are scene means. Network totals and scene optimization variables follow the definitions and timing scopes in the text.}
\label{tab:exp_runtime_all37}
\label{tab:exp_parameters_all37}
\end{table*}

\begin{table*}[!t]
\begin{minipage}[t]{0.48\textwidth}
\centering
\small
\setlength{\tabcolsep}{3pt}
\begin{tabular}{l|rr|rr}
\toprule
& \multicolumn{2}{c|}{DL3DV} & \multicolumn{2}{c}{Mip-NeRF 360} \\
\cmidrule(lr){2-3}\cmidrule(lr){4-5}
Method & Enhancement & Transfer & Enhancement & Transfer \\
\midrule
BilaRF & 73.3 & 70.2 & 218.4 & 226.2 \\
\rowcolor{retouchpale}
\method & \textcolor{retouchdeep}{\textbf{5.6}} & \textcolor{retouchdeep}{\textbf{5.0}} & \textcolor{retouchdeep}{\textbf{32.8}} & \textcolor{retouchdeep}{\textbf{32.9}} \\
\bottomrule
\end{tabular}
\caption{Median preparation-inclusive delivery time on an RTX 5090 (seconds).}
\label{tab:delivery_time}
\end{minipage}\hfill
\begin{minipage}[t]{0.48\textwidth}
\centering
\footnotesize
\setlength{\tabcolsep}{4pt}
\begin{tabular}{l|rr|rr}
\toprule
& \multicolumn{2}{c|}{DL3DV} & \multicolumn{2}{c}{Mip-NeRF 360} \\
\cmidrule(lr){2-3}\cmidrule(lr){4-5}
Component & Grid LUT & GLUT & Grid LUT & GLUT \\
\midrule
Fitting & 0.123 & 20.234 & 0.124 & 23.680 \\
Blended color context & 0.629 & 0.629 & 2.812 & 2.812 \\
Analytic color update & 0.095 & 0.776 & 0.234 & 2.961 \\
Color refinement & 0.149 & 0.560 & 0.468 & 2.328 \\
\midrule
Total update & 0.256 & 1.335 & 0.717 & 5.289 \\
\bottomrule
\end{tabular}
\caption{Median component times for reference-guided color transfer on an RTX 5090 (seconds). Total update sums analytic color update and blend-aware color refinement per scene.}
\label{tab:operator_timing}
\end{minipage}
\end{table*}

\subsection{Rendering Requirements}
\label{app:rendering_requirements}

We compare rendering requirements during asset preparation and editing.
Starting from a pretrained 3DGS asset, image-conditioned enhancement baselines render source views to construct their targets.
BilaRF~\cite{wang2024bilarf}, SAAD~\cite{saad}, and Instant Colorization~\cite{instantcolorization} also use rendered supervision during editing. ReparamGS~\cite{reparamgs}, FPGS~\cite{fpgs}, and VIRGi~\cite{virgi} render views for scene-specific preparation.
ViP3DE~\cite{chen2026vip3de} and Omni-3DEdit~\cite{chen2026omni3dedit} propagate edits through source images, while VCG~\cite{video_color_grading} uses an input-view sequence to predict its LUT.
In contrast, \method predicts the scene-level color operator with its aesthetic enhancement network and applies the render-free color update directly to the asset.
Reference-guided color transfer renders the original reference view once for paired operator fitting, followed by the same render-free color update.

\subsection{Computational Cost}
\label{app:cost_inventory}
\label{app:runtime}

\noindent{\textbf{Execution Time and Parameter Counts.}}
\Cref{tab:exp_runtime_all37} summarizes mean execution time, network parameters, and scene-specific optimization variables over all 37 scenes.

Execution time in \cref{tab:exp_main,tab:exp_runtime_all37} includes asset loading, editing, and native-output export, measured with synchronized GPU boundaries, with the method-specific scopes below.
Image-target generation, evaluation, and preview rendering are timed separately.
ReparamGS, FPGS, and VIRGi include scene preparation for enhancement and reuse it for transfer.
All methods are run on a single NVIDIA RTX 5090 GPU with 32 GB memory.

ViP3DE enhancement includes input preparation and editing. Its transfer interval includes model loading, propagation, asset fitting, validation, and export, with source rendering and input preparation performed beforehand.
Omni-3DEdit includes input preparation and two image-propagation batches, with source rendering and global model initialization performed beforehand. First-use initialization is included when it occurs.
GLUT times cover fitting, context preparation, and color update.

Network parameter counts include each instantiated model component once, including frozen modules and the NamedCurves enhancer used by image-conditioned enhancement baselines.
For GLUT, they include the 716 parameters of its Gaussian LUT representation.
Scene variables count quantities optimized or solved during preparation and editing, averaged across scenes when their size depends on the Gaussian population.
Prep. vars. and Edit vars. distinguish one-time scene preparation from optimization for a particular edit.
These counts describe different aspects of a method: a scene-trained network contributes to both, while a direct color solve contributes only scene variables.
Methods with zero network parameters use analytic color updates or direct optimization of scene-specific color variables.
Network parameter counts exclude fixed Gaussian storage, optimizer state, and evaluation models.
The grid LUT used for transfer has 14,739 fitted values. Aesthetic enhancement predicts its LUT with no per-scene optimization.

\noindent{\textbf{Preparation-Inclusive Delivery.}}
\Cref{tab:delivery_time} reports median delivery times on the RTX 5090, including asset loading, scene preparation, editing, and export.
When preparation is reused, its original computation time is included to represent delivery of the first edit.

\noindent{\textbf{Color-Update Components.}}
Component times in \cref{tab:operator_timing} measure fitting, context preparation, and color updates separately.
Grid fitting uses a GPU-synchronized solver interval.
GLUT fitting includes optimization, validation, and parameter export.
Total update sums the analytic color update and blend-aware color refinement, while delivery times additionally include asset loading and export.

\begin{figure*}[p]
\centering
\includegraphics[width=\textwidth]{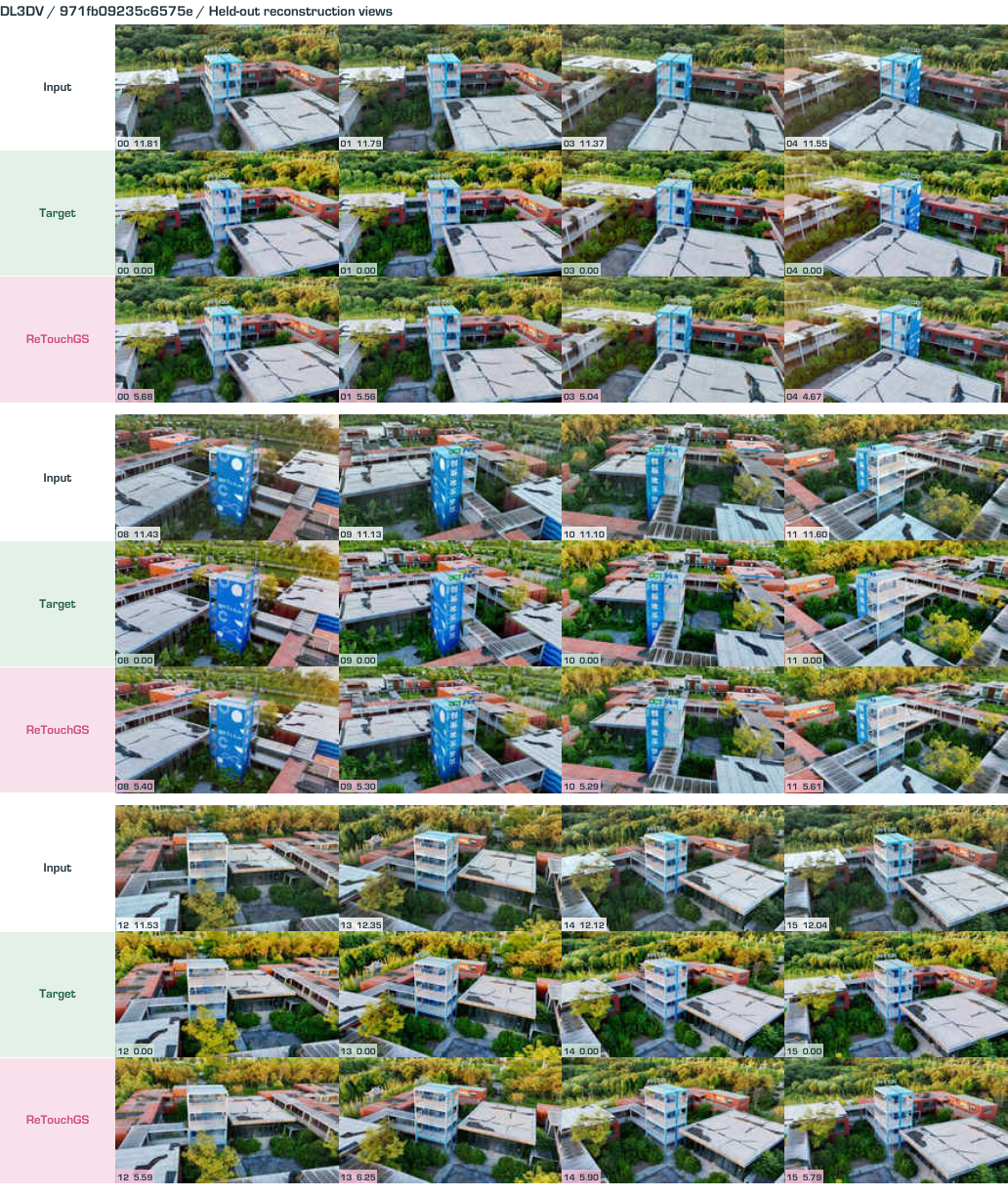}
\caption{Multi-view aesthetic color enhancement from one asset at held-out cameras. Columns align Input, Target, and \method, with view indices and crop \dE$\downarrow$ against Target.}
\label{fig:app_enhancement_multiview}
\end{figure*}

\begin{figure*}[p]
\ContinuedFloat
\centering
\includegraphics[width=\textwidth]{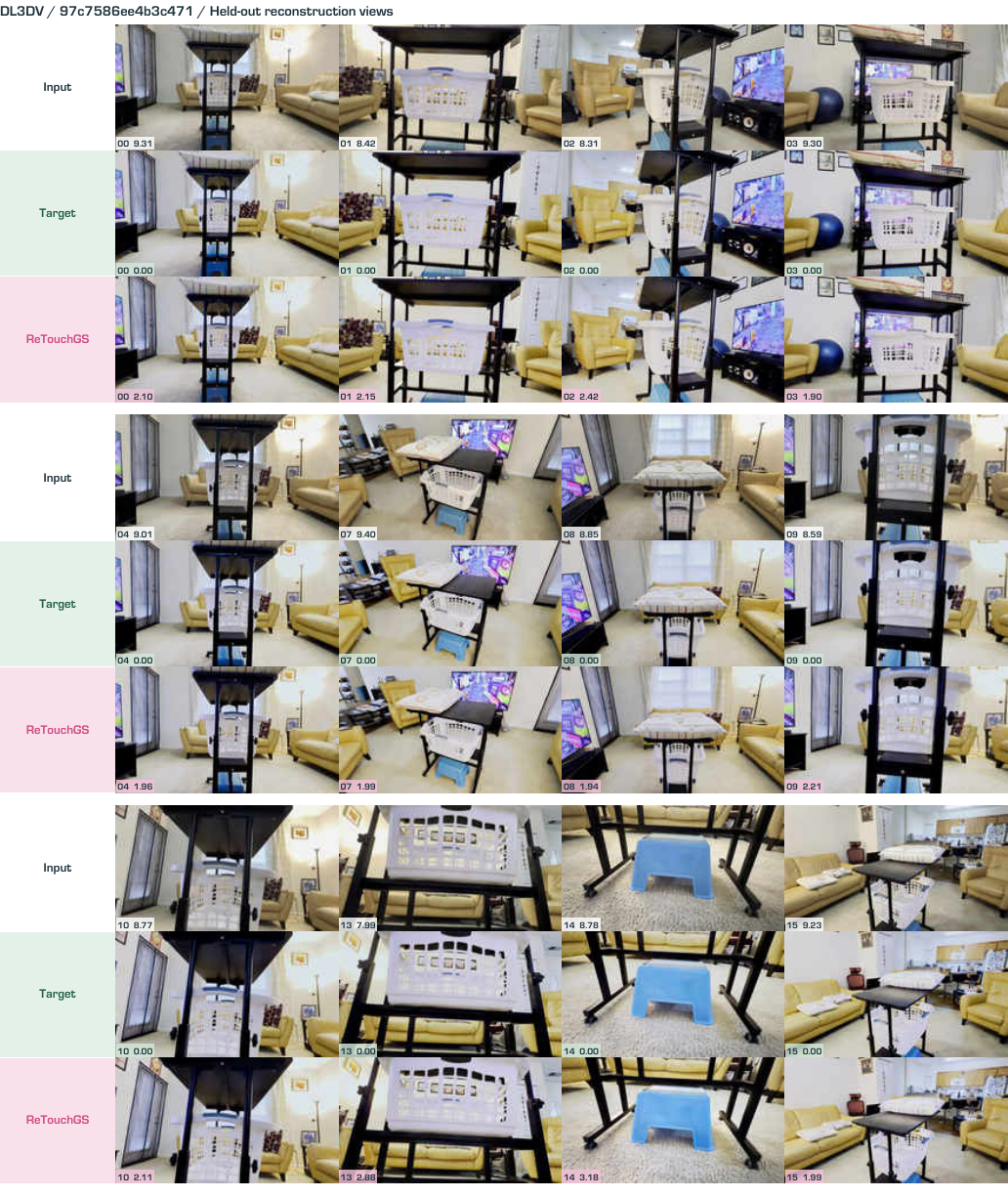}
\caption[]{Multi-view aesthetic color enhancement on a second scene, with the same layout and metrics.}
\label{fig:app_enhancement_multiview_continued}
\end{figure*}

\begin{figure*}[p]
\centering
\includegraphics[width=\textwidth]{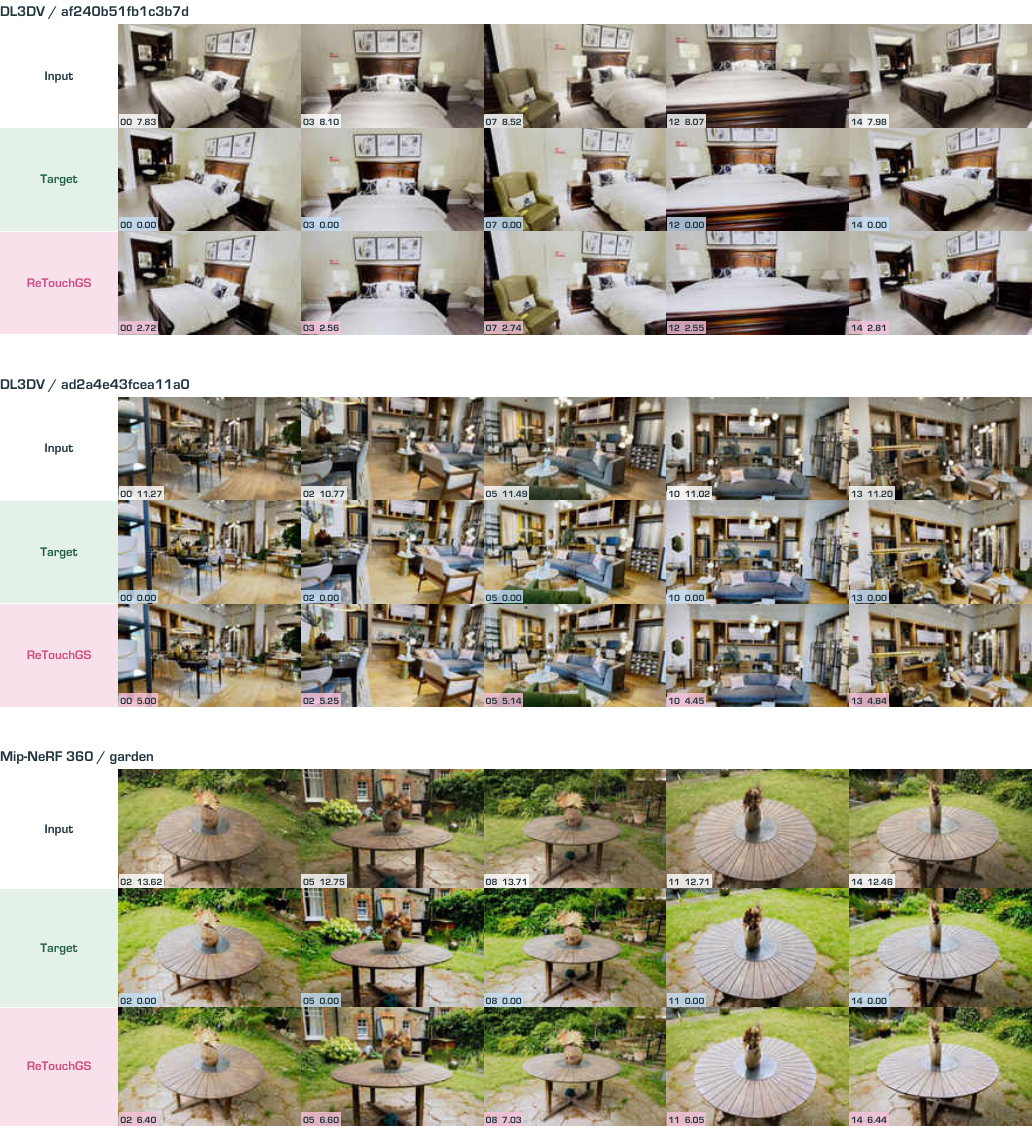}
\caption{Aligned multi-view aesthetic color enhancement. Each scene shows Input, Target, and \method at five held-out cameras. All outputs for a scene come from one enhanced asset. Corner labels give the view index and mean pixelwise \dE$\downarrow$ over the displayed region.}
\label{fig:app_aligned_multiview}
\end{figure*}

\begin{figure*}[p]
\ContinuedFloat
\centering
\includegraphics[width=\textwidth]{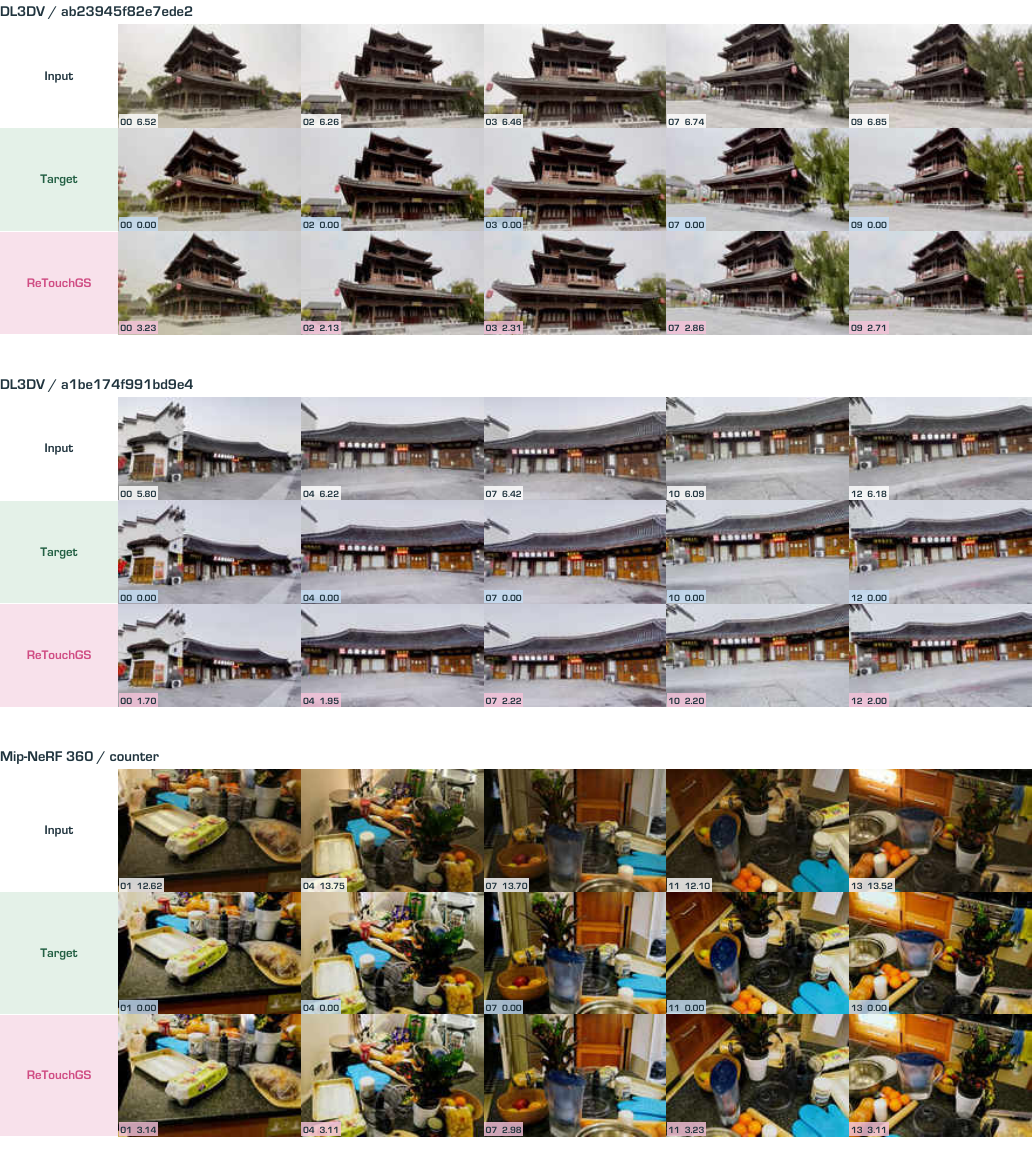}
\caption[]{Aligned multi-view reference-guided color transfer with the reference-fitted grid LUT. Each scene shows Input, Target, and \method at five held-out cameras. All outputs for a scene come from one edited asset. Corner labels give the view index and mean pixelwise \dE$\downarrow$ over the displayed region.}
\label{fig:app_aligned_multiview_transfer}
\end{figure*}

\end{document}